\documentclass[3p, preprint, 10pt, twocolumn]{elsarticle}

\makeatletter\if@twocolumn\PassOptionsToPackage{switch}{lineno}\else\fi\makeatother

\usepackage{subcaption}
\usepackage[ruled,vlined]{algorithm2e}
\usepackage[toc,page]{appendix}
\usepackage{tabulary,xcolor}
\usepackage{amsfonts,amsmath,amssymb}
\usepackage[T1]{fontenc}
\makeatletter
\let\save@ps@pprintTitle\ps@pprintTitle
\def\ps@pprintTitle{\save@ps@pprintTitle\gdef\@oddfoot{\footnotesize\itshape \null\hfill\today}}
\def\hlinewd#1{%
  \noalign{\ifnum0=`}\fi\hrule \@height #1%
  \futurelet\reserved@a\@xhline}

\AtBeginDocument{\ifNAT@numbers \biboptions{sort&compress}\fi}
\makeatother

\usepackage{ifluatex}
\ifluatex
\usepackage{fontspec}
\defaultfontfeatures{Ligatures=TeX}
\usepackage[]{unicode-math}
\unimathsetup{math-style=TeX}
\else 
\usepackage[utf8]{inputenc}
\fi 
\ifluatex\else\usepackage{stmaryrd}\fi

\usepackage{url,multirow,morefloats,floatflt,cancel,tfrupee}
\makeatletter

\AtBeginDocument{\@ifpackageloaded{textcomp}{}{\usepackage{textcomp}}}
\makeatother
\usepackage{colortbl}
\usepackage{xcolor}
\usepackage{pifont}
\usepackage[nointegrals]{wasysym}
\makeatletter

\def\mcWidth#1{\csname TY@F#1\endcsname+\tabcolsep}

\def\cAlignHack{\rightskip\@flushglue\leftskip\@flushglue\parindent\z@\parfillskip\z@skip}
\def\rAlignHack{\rightskip\z@skip\leftskip\@flushglue \parindent\z@\parfillskip\z@skip}

\@ifundefined{etal}{}{}

\usepackage{ifxetex}
\ifxetex\else\if@twocolumn\@ifpackageloaded{stfloats}{}{\usepackage{dblfloatfix}}\fi\fi

\AtBeginDocument{
\expandafter\ifx\csname eqalign\endcsname\relax
\def\eqalign#1{\null\vcenter{\def\\{\cr}\openup\jot\m@th
  \ialign{\strut$\displaystyle{##}$\hfil&$\displaystyle{{}##}$\hfil
      \crcr#1\crcr}}\,}
\fi
}

\AtBeginDocument{%
  \@ifpackageloaded{endfloat}%
   {\renewcommand\efloat@iwrite[1]{\immediate\expandafter\protected@write\csname efloat@post#1\endcsname{}}}{\newif\ifefloat@tables}%
}%

\def\BreakURLText#1{\@tfor\brk@tempa:=#1\do{\brk@tempa\hskip0pt}}
\let\lt=<
\let\gt=>
\def\processVert{\ifmmode|\else\textbar\fi}

\@ifundefined{subparagraph}{
\def\subparagraph{\@startsection{paragraph}{5}{2\parindent}{0ex plus 0.1ex minus 0.1ex}%
{0ex}{\normalfont\small\itshape}}%
}{}

\newcommand\role[1]{\unskip}
\newcommand\aucollab[1]{\unskip}
  
\@ifundefined{tsGraphicsScaleX}{\gdef\tsGraphicsScaleX{1}}{}
\@ifundefined{tsGraphicsScaleY}{\gdef\tsGraphicsScaleY{.9}}{}
\def\checkGraphicsWidth{\ifdim\Gin@nat@width>\linewidth
	\tsGraphicsScaleX\linewidth\else\Gin@nat@width\fi}

\def\checkGraphicsHeight{\ifdim\Gin@nat@height>.9\textheight
	\tsGraphicsScaleY\textheight\else\Gin@nat@height\fi}

\def\fixFloatSize#1{}
\let\ts@includegraphics\includegraphics

\def\inlinegraphic[#1]#2{{\edef\@tempa{#1}\edef\baseline@shift{\ifx\@tempa\@empty0\else#1\fi}\edef\tempZ{\the\numexpr(\numexpr(\baseline@shift*\f@size/100))}\protect\raisebox{\tempZ pt}{\ts@includegraphics{#2}}}}

\AtBeginDocument{\def\includegraphics{\@ifnextchar[{\ts@includegraphics}{\ts@includegraphics[width=\checkGraphicsWidth,height=\checkGraphicsHeight,keepaspectratio]}}}

\DeclareMathAlphabet{\mathpzc}{OT1}{pzc}{m}{it}

\def\URL#1#2{\@ifundefined{href}{#2}{\href{#1}{#2}}}

\def\UrlOrds{\do\*\do\-\do\~\do\'\do\"\do\-}%
\g@addto@macro{\UrlBreaks}{\UrlOrds}

\edef\fntEncoding{\f@encoding}

\makeatother

\newif\ifmultipleabstract\multipleabstractfalse%
\begin{document}

\begin{frontmatter}
	
\title{Bidirectional Interaction between Visual and Motor Generative Models using Predictive Coding and Active Inference}
    
\author[]{Louis Annabi\corref{c-8d7e5a86f115}}
\ead{louis.annabi@ensea.fr}\cortext[c-8d7e5a86f115]{Corresponding author.}
\author[]{Alexandre Pitti}
\ead{alexandre.pitti@ensea.fr}
\author[]{Mathias Quoy}
\ead{mathias.quoy@ensea.fr}
    
\address{ETIS UMR 8051, CY University, ENSEA, CNRS}

\begin{abstract}

In this work, we build upon the Active Inference (AIF) and Predictive Coding (PC) frameworks to propose a neural architecture comprising a generative model for sensory prediction, and a distinct generative model for motor trajectories. We highlight how sequences of sensory predictions can act as rails guiding learning, control and online adaptation of motor trajectories. We furthermore inquire the effects of bidirectional interactions between the motor and the visual modules. The architecture is tested on the control of a simulated robotic arm learning to reproduce handwritten letters.

\end{abstract}


\tnotetext[t1]{This work was funded by the CY Cergy-Paris University Foundation (Facebook grant) and partially by Labex MME-DII, France (ANR11-LBX-0023-01).}

\end{frontmatter}
    
\section{Introduction}

In this work, we tackle the problem of motor sequence learning for an embodied agent. A wide range of approaches have been proposed to model sequential data, using various types of neural architectures (Recurrent Neural Networks (RNNs), Long Short-Term Memories (LSTMs) \cite{Hochreiter1997}, Restricted Boltzmann Machines (RBMs) \cite{Sutskever2009}) and various learning strategies (backpropagation through time (BPTT), Real-Time Recurrent Learning (RTRL) \cite{Williams1989}, Reservoir Computing (RC) \cite{Verstraeten2007, Lukosevicius2009}).

In an embodied simulation, the agent continuously performs motor commands, or actions, that influence its environment, and continuously perceives information about the state of its environment through sensory observations. Given a data set of motor trajectories, one could train a generative model using the methods cited above to learn a repertoire of motor trajectories. However, considering the constraint of embodiment, we cannot assume the availability of supervision in the agent's motor space. Instead, supervision can be available in the shape of desired sensory observations, for instance provided by a teaching agent. In the case of handwriting, these desired sensory observations are visual observations of the target letters. In reinforcement learning, the preference for certain sensory states is modeled by assigning rewards to the desired states, and the agent learns a behavioral policy maximizing its expected return (sum of rewards) over time. Alternatively, Active Inference (AIF) \cite{Friston2009a, Friston2016}, derived from the Free Energy Principle (FEP) \cite{Friston2006, Buckley2017}, proposes to see acting as a way of minimizing surprise, by choosing to perform actions that will produce sensory observations that are probable under the agent's generative model. In other words, to perform actions leading to desired sensory states, the agent must learn a generative model naturally inclined towards predicting the desired sensory states, and perform the actions that fulfill these predictions.

In our work, we propose to learn a generative model of the trajectories in the sensory space, that we exploit in order to guide the generation of motor trajectories using AIF. We present a neural network architecture based on two distinct RNNs generating sequences in the sensory and motor spaces. We apply our approach to the problem of learning joint angle motor trajectories for handwriting with a simulated 3 DoF articulated arm. Target trajectories are provided in the visual space, as sequences of 2D pen positions, and the agent has to learn a repertoire of corresponding motor trajectories.

Our complete model incorporates different components that can be trained in subsequent stages. First, AIF requires a forward model, that is, a model of how the agent's actions affect its observations. Such models can be learned in early developmental stages via a random interaction with the environment, also called motor babbling. This approach has been widely used to learn the relations between motor commands and sensory observations without any external supervision (e.g. \cite{Mochizuki2013, Sasaki2019}). Second, AIF requires a generative model for trajectories in the visual space. This model can be learned from the supervision provided by the target trajectories. In this work we propose to implement this generative model with an RNN design in line with the FEP and Predictive Coding (PC) \cite{Rao1999, Clark2013}, inspired from \cite{Ororbia2020, Friston2009c}. Finally, our architecture will make use of these two subsystems to train a second generative model in the motor space using AIF, which we implement using another instance of the PC inspired RNN model.

Several works advocate for the relevance of randomly connected RNNs as a computational model for cortical networks \cite{Wang2010, Hoerzer2012, Mannella2015}. In particular, \cite{Mannella2015} suggests RC as a candidate approach to generate movements as neural trajectories in the motor cortex. However, the authors propose to train these cortical networks through a supervised learning scheme, which would need target values in the motor space. 

Instead, our approach relates to the internal model theory, suggesting that efferent copies of motor commands in the brain are provided as inputs to an internal forward model predicting the sensory outcomes of performed actions \cite{Shadmehr2010}. The interesting feature brought by AIF is that, in contrast with control theory where the heavy lifting is done by the inverse models, the reciprocal top-down and bottom-up information passing scheme allows to infer proper actions using an error signal between sensory predictions and predicted outcomes of actions. These types of internal models are thought to be encoded in the intraparietal sulcus and superior parietal lobule regions of the posterior parietal cortex, for reaching and grasping movements \cite{Creemregehr2009}, as well as drawing and handwriting \cite{Planton2017}.

On a higher level of abstraction, motor cognition (planning, decision making) involves other brain structures such as the cerebellum and the prefrontal cortex, for the prediction of outcomes \cite{Pezzulo2016, Mushiake2006, Botvinick2009}, and the basal ganglia, for the selection of action policies \cite{Botvinick2009, Mannella2015}. In \cite{Friston2016}, AIF is proposed as a candidate model for goal-directed behaviour relating to the brain structures cited above.

While our work does not aim at providing a computational model of brain functions, the theories emerging from research in computational neurosciences still serve as an inspiration to build functional models to be integrated on robotic platforms. The contribution brought by this work is two-fold. First, we show how AIF makes it possible to learn a repertoire of motor trajectories without requiring supervision in the motor space, inverse model learning, or BPTT. Second, we show that the dynamic interactions between the sensory and motor generative models, implemented with PC, provide relevant properties for motor control : robustness to external perturbations, adaptation to variations of size or orientation of the target trajectory, intermittent control according to a precision threshold.

We will first present related works in the fields of RNNs and models for handwriting. Then, we will describe our architecture, before reporting and analyzing the results obtained in several experimental setups.

\section{Related work}

\subsection{Recurrent Neural Networks}

\begin{figure*}[!ht]
    \centering
    \begin{subfigure}{0.087\textwidth}
        \centering
        \includegraphics[width=\textwidth]{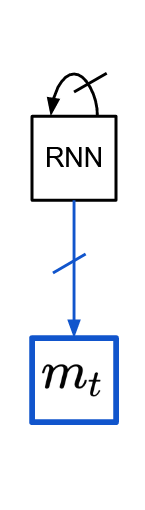}
        \caption{}
        \label{fig:supervised_model}
    \end{subfigure}
    ~
    \begin{subfigure}{0.143\textwidth}
        \centering
        \includegraphics[width=\textwidth]{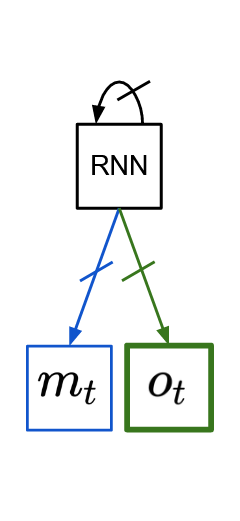}
        \caption{}
        \label{fig:habituated_trajectories_model}
    \end{subfigure}
    ~
    \begin{subfigure}{0.33\textwidth}
        \centering
        \includegraphics[width=\textwidth]{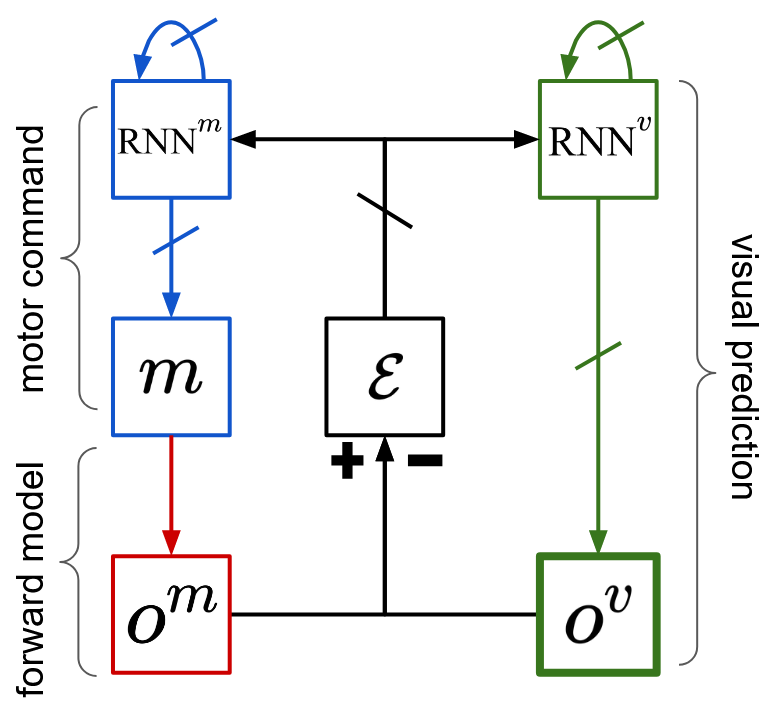}
        \caption{}
        \label{fig:model_overview}
    \end{subfigure}%
    ~
    \begin{subfigure}{0.38\textwidth}
        \centering
        \includegraphics[width=\textwidth]{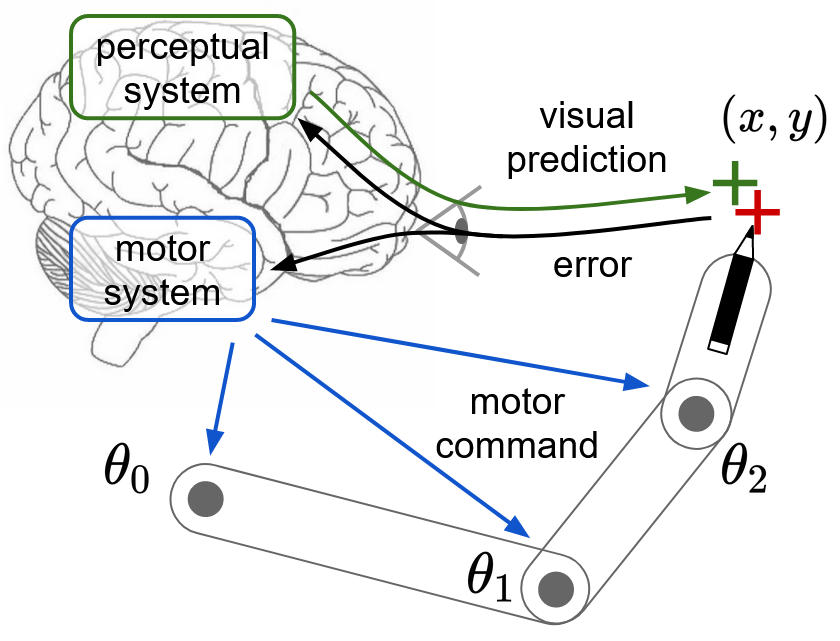}
        \caption{}
        \label{fig:brain_arm}
    \end{subfigure}%
    \caption{Different related approaches to learn motor trajectories. Synaptic weights are represented by marks on the arrows. Variables for which target values are provided are contained in bold squares. \textbf{a}: Supervised learning of a generative model for motor trajectories. \textbf{b}: Learning of a generative model for sensorimotor trajectories with supervision in the sensory space. \textbf{c}: Learning of separate generative models for motor trajectories and sensory predictions. \textbf{d}: Connection between our experimental setup and the model displayed in \textbf{c}. $\mathbf{m} = (\theta_0, \theta_1, \theta_2)$ and $\mathbf{o} = (x,y)$.}
    \label{fig:full_model}
\end{figure*}

Recurrent neural networks (RNNs) form a category of models that can be used for sequence generation and recognition tasks. An RNN can be seen as a dynamical system influenced by inputs. At each time step, it updates its state based on its past state and its current input. Additionally, the RNN can include a readout layer, decoding the RNN state sequence into an output sequence. There exist several approaches to train RNNs. The input weights, recurrent weights, and output weights of RNNs can be learned through backpropagation through time. However, it has been proved that this optimization method can give rise to exponentially decaying or exploding gradients \cite{Pascanu2012}, thus making the learning either slow or unstable. Several solutions have been proposed to address this issue, such as hessian free optimization \cite{Martens2010, Martens2011}, or gating mechanisms for capturing long-term dependencies \cite{Hochreiter1997, Graves2013}. Additionally, the BPTT algorithm is arguably impossible to be implemented in the brain, the main reason being the non-locality of the information used for gradient computations.

Other approaches to RNN training providing more biologically plausible mechanisms for learning have been researched. Completely avoiding the problem of learning recurrent weights is a family of approaches that emerged in parallel from the fields of computational neurosciences \cite{Maass2002} and machine learning \cite{Jaeger2001} and has been labeled later as Reservoir Computing (RC) \cite{Lukosevicius2009}. By carefully initializing the recurrent weights of the RNN, RC methods completely put aside the problem of learning these parameters. The recurrent connections are set in order for the RNN to exhibit rich non-linear and sometimes self-sustained dynamics, that are decoded by a learned readout layer. RC has been used for generation tasks \cite{Laje2013} and recognition tasks \cite{Triefenbach2010}. \cite{Vlachas2020} provides an experimental comparison of RC and BPTT approaches for different sequential tasks.

Finally, learning methods inspired from the Predictive Coding (PC) theory have been proposed as an alternative for backpropagation \cite{Whittington2017, Millidge2020}. PC \cite{Rao1999, Clark2013} is a theory of brain function in which the brain is constantly and hierarchically generating top-down predictions about its sensory states, and updating its internal states based on a bottom-up error signal originating from the sensory level. This view is to some extent supported by neurophysiological data \cite{Walsh2020}, and aligns nicely with the Bayesian brain hypothesis assuming that the brain implements a form of Bayesian inference \cite{Dayan1995, Knill2004}.

In neural implementations of PC, the generative model is intertwined with error neurons that propagate the information in a bottom-up manner through the hierarchy. An online estimation of the error at each level of the generative model makes it possible to learn the model parameters, and infer the hidden states, using only local update rules. Following this approach, \cite{Ororbia2020} proposed a recurrent neural network architecture that yields state of the art performance on the bouncing MNIST task. In \cite{Friston2009c}, the authors propose a PC architecture where the generative model incorporates hidden causes in addition to hidden states. Contrary to hidden states that are embedded with dynamics, hidden causes are static variables in the generative model. These variables also differ from model parameters since they can be dynamically updated through online inference (using the bottom-up error circuitry). Their model conditions the dynamics of the hidden states using these hidden causes. They show that their approach allows synthetic birds to recognize and categorize bird songs, using a generative model based on a cascade of Lorenz attractors. 

\subsection{Handwriting}

Handwriting and drawing require fundamental cognitive abilities involving visual and motor skills, that make it possible to translate abstract representations into visuomotor trajectories. Many computational models for handwriting and drawing simply consider this problem as that of generating sequences of pen positions. This category of implementations is represented in figure \ref{fig:supervised_model} and can use the different approaches for RNN training cited before. \cite{Graves2013, Ha2017} propose LSTMs with mixture density outputs trained with BPTT, \cite{Mannella2015} exposes a neurocomputational model for the selection of motor sequences implemented with RC, and \cite{Philippsen2020} uses a computational model based on PC and Bayesian inference. 

However, in an embodied simulation, it is unclear how direct supervision in the motor space could occur. Instead, in the proposed task, supervision takes place in the sensory space, and motor sequences are generated in order to reproduce target sequences of visual observations. Addressing this challenge are the approaches presented in \cite{Mochizuki2013, Sasaki2019}. In these works, the authors avoid the difficult problem of inverse model learning by training an RNN to jointly generate sensory and motor trajectories, as represented in figure \ref{fig:habituated_trajectories_model}. Their generative model is first trained on random pairs of motor commands and visual observations obtained with motor babbling. Later, they infer an initial RNN state that properly predicts a target visual trajectory. This initial RNN state is used to predict and perform the motor trajectory. Finally, the model weights are tuned on the performed trajectories to maintain the coherence between predicted visual observations and motor commands. The relation between motor and visual trajectories is here embedded within the generative model, but is only accurate on the learned trajectories, that can be referred to as "habituated trajectories".

Another way to approach this constraint is to take inspiration from the FEP and AIF \cite{Friston2009a, Friston2016}. The starting point of the FEP is that agents maintain homeostasis with their environments, which is made possible by minimizing surprise through both action and perception. Free energy intervenes as an upper-bound on surprise, that can be computed and thus optimized more efficiently. FEP applied to action casts motor control and decision making into the same process of surprise minimization. The agent is endowed with a generative model predicting sensory observations that can be naturally biased towards desired observations. Minimizing surprise through action corresponds to the inference of an action that will cause sensory states probable under this generative model.

This framework has gained popularity and there are many implementations of AIF in the litterature. \cite{Friston2009a} shows how AIF can reproduce behavioral policies obtained using reinforcement learning and dynamic programming methods. In \cite{PioLopez2016}, the authors implement AIF to perform reaching with a 7-DoF simulated arm. \cite{Oliver2019} proposes a robotic implementation of AIF on an iCub robot performing reaching and active head object tracking. See \cite{Ciria2021} for a recent and more complete review about models implementing AIF.

In previous works, we applied the FEP to the control of long range neural synchrony in recurrent spiking networks \cite{Pitti2017}. The proposed model was able to generate very long and precise spatio-temporal sequences, using a random search algorithm to optimize free energy. In \cite{Annabi2020}, we proposed a self-supervised algorithm in line with AIF to learn repertoires of motor primitives for a simulated robotic arm. Free energy minimization was used to regress adequate initial hidden states of the RNN generating motor trajectories. In comparison, the work we present here allows for an online control throughout the whole motor trajectory.

Our approach stands out by including a second separate generative model for motor trajectories, as represented in figure \ref{fig:model_overview}. Contrary to direct implementations of active inference, it uses the gradient descent on surprise only as a mechanism to update its prior belief on the motor command. This approach makes it possible to simulate a bidirectional interaction between action and perception generative models. This is consistent with studies from developmental psychology suggesting that reading training improves handwriting skills \cite{Vinter2010}, and reversely, that handwriting training improves letter recognition scores \cite{Longcamp2005}.

\section{Methods}

In this section we present our model for motor trajectories learning. We first describe the overall structure, before detailing each individual component.

\subsection{Architecture}

\begin{figure*}[!ht]
    \centering
    \includegraphics[width=0.75\textwidth]{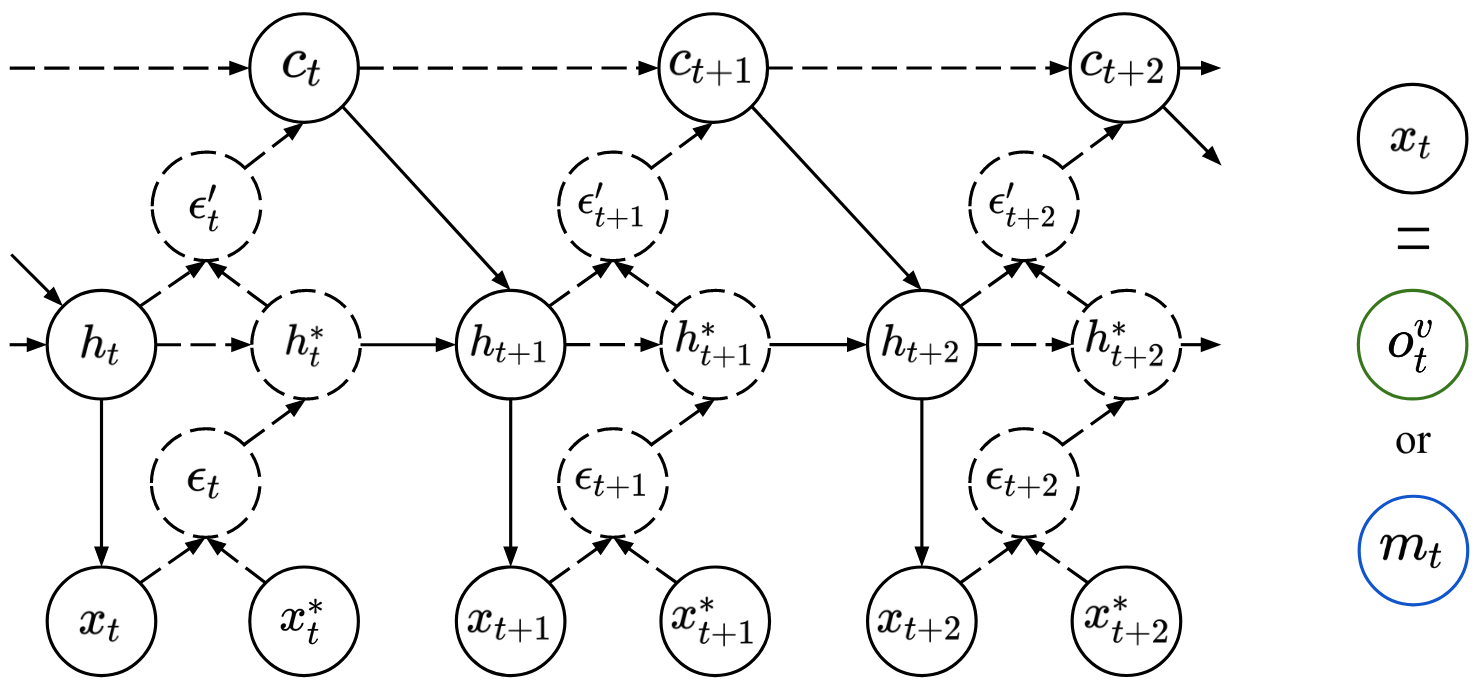}
    \caption{Recurrent neural network model used to model the sequence generation in the sensory and motor spaces. Predictions are denoted $\mathbf{x}$ and target values are denoted $\mathbf{x}^*$. The middle layer corresponds to the hidden state of the generative model, denoted $\mathbf{h}$. The upper layer corresponds to the hidden causes of the generative model, denoted $\mathbf{c}$. Dashed arrows represent the bottom-up pathway for online inference of hidden states and hidden causes. The variables $\mathbf{\epsilon}$ and $\mathbf{\epsilon}'$ correspond to the prediction errors at the different layers of the generative model. They are used to infer an a posteriori estimate of the hidden states (denoted $\mathbf{h}^*$) and hidden causes variables.}
    \label{fig:perceptual_model}
\end{figure*}

Our embodied agent perceives information from its environment via visual observations that we denote $\mathbf{o}_t$, and can influence the state of the environment $\mathbf{s}_t$ via motor commands, denoted $\mathbf{m}_t$. We separate motor and visual pathways into two distinct dynamical systems interacting with each other only via a control mechanism minimizing prediction error on the visual level. Figure \ref{fig:model_overview} displays an overview of our computational model for motor sequence learning.

In early stage of its development, we assume that our agent acquires a suitable forward model of its environment, denoted $f$, predicting its visual observation based on its motor command and the previous state of the environment: $\mathbf{o}_t = f(\mathbf{s}_{t-1}, \mathbf{m}_t)$. Since our work do not focus on the learning of such a model, we omitted the dependency according to $\mathbf{s}_{t-1}$ to simplify the graph in figure \ref{fig:model_overview}.

Our agent's training is composed of the following stages : 
\begin{itemize}
    \item Learning of a visual generative model, predicting trajectories in the visual space.
    \item Learning of a motor generative model according to the visual generative model and forward model.
\end{itemize}

\subsection{Predictive coding recurrent neural network}
\label{sec:predictive_coding_rnn}

Figure \ref{fig:perceptual_model} represents the RNN model we use for the prediction and learning of the trajectories in the visual and motor space. This RNN implementation combines several ideas from \cite{Ororbia2020, Taylor2009, Friston2009c}.

Taking inspiration from \cite{Friston2009c}, we make a distinction between hidden states  $\mathbf{h}$ and hidden causes $\mathbf{c}$ in our generative model. The hidden state is a dynamic variable, in our case corresponding to the internal state of an RNN. The hidden causes is a static variable that can influence the dynamics of the hidden state variable. To model this influence, we use a gain-field network (e.g. \cite{Abrossimoff2018}) implemented as a three-way weight tensor $\mathbf{W_{rec}}$ for the recurrent connections in the RNN. We interpret this tensor as a basis of size $p$ of recurrent weights matrices of shape $(n \times n)$. Here, $n$ and $p$ stand respectively for the dimensions of the hidden state and hidden causes. Multiplying this tensor with the hidden causes vector can thus be seen as computing a recurrent weight matrix from this basis using the hidden causes as coordinates. It ensues that different hidden causes will provide different hidden state dynamics.

To avoid scaling issues when dealing with three-way tensors, \cite{Taylor2009} proposes to factor it into three matrices, such that for all $i, j, k$, $\mathbf{W}_\mathbf{rec}^{ijk} = \sum_{l<d}\mathbf{W}_\mathbf{p}^{il} \cdot \mathbf{W}_\mathbf{f}^{jl} \cdot \mathbf{W}_\mathbf{c}^{kl}$. The model can scale better to large hidden state and hidden causes dimensions with this factorization, since the dimension $d$ can be adjusted to control the number of model parameters. In our experiments, we will always use $d=n/2$. The three matrices $\mathbf{W_p}$, $\mathbf{W_f}$ and $\mathbf{W_c}$ respectively model the interaction of the past hidden state, future hidden state, and hidden causes with the factor dimension.

Building upon the RNN model proposed by \cite{Ororbia2020}, our model is able to learn to generate supervised trajectories using only local learning rules, and thus not requiring backpropagation of gradients through time. Contrary to purely top-down generative models, only capable of \textit{prediction}, our model is also capable of performing \textit{inference}. This is achieved by taking inspiration from the PC theory: the RNN is augmented with error neurons, denoted $\mathbf{\epsilon}$ and $\mathbf{\epsilon'}$, measuring the shift between the predicted and the target values at each layer. These error neurons are mapped onto their upper layer via feedback connections, to update either the hidden state $\mathbf{h}$ or the hidden causes $\mathbf{c}$ of the RNN, thus performing a form of Bayesian inference by updating beliefs based on new evidence.

The following equations describe all the computations occurring for one time step in the RNN, in the top-down pathway, for \textit{prediction}:
\begin{equation}
    \begin{aligned}
    \mathbf{h}_t &= (1-\frac{1}{\tau}) \mathbf{h}^*_{t-1} + \frac{1}{\tau} \mathbf{W_f} \\
    & \cdot ((\mathbf{W_c}^\intercal \cdot \mathbf{c}_{t-1})(\mathbf{W_p}^\intercal \cdot \tanh(\mathbf{h}_{t-1}^*))) \\
    \end{aligned}
\end{equation}
\begin{equation}
    \mathbf{x}_t = \mathbf{W_{out}} \cdot \tanh(\mathbf{h}_t)
\end{equation}

And in the bottom-up pathway, for \textit{inference}:
\begin{align}
    \mathbf{\epsilon}_t &= \mathbf{x}_t - \mathbf{x}_t^* \\
    \mathbf{h}_t^* &= \mathbf{h}_t - \alpha_h \mathbf{W_{out}}^\intercal \cdot \mathbf{\epsilon_t} \label{eq:hidden_state_update}\\
    \mathbf{\epsilon}_t' &= \mathbf{h}_t - \mathbf{h}_t^*
\end{align}
\begin{equation}
    \begin{aligned}
        \mathbf{c}_t &= \mathbf{c}_{t-1} - \alpha_c \mathbf{W_c} \\
        & \cdot ((\mathbf{W_f}^\intercal \cdot \mathbf{\epsilon}_t')(\mathbf{W_p}^\intercal \cdot \tanh(\mathbf{h}_{t-1}^*)))  \label{eq:hidden_causes_update}
    \end{aligned}
\end{equation}

We introduced a time constant $\tau$ influencing the pace of the hidden state dynamics, as well as parameters $\alpha_h$ and $\alpha_c$, controlling for the weight of the incoming bottom-up information in the a posteriori estimation of the hidden states and hidden causes. We can discuss the choice of reusing the weights of the top-down pathway in the bottom-up pathway. Other approaches consider using random feedback weights \cite{Lillicrap2016, Nokland2016}, or new sets of weights that can be learned \cite{Ororbia2020}. Reusing the top-down weights provides a simple solution as the equations \ref{eq:hidden_state_update} and \ref{eq:hidden_causes_update} correspond to gradient descent updates on the hidden states and hidden causes. Compared to random weights, it led in our experiments to better accuracy during online inference, and faster convergence during learning (results not shown).

Learning is performed in this RNN using only local online gradient descent rules. The output weights $\mathbf{W_{out}}$ are updated in order to minimize the prediction error on the visual level, and the recurrent weights $\mathbf{W_p}$, $\mathbf{W_f}$ and $\mathbf{W_c}$ are updated in order to minimize the error between the prior hidden state $\mathbf{h}$ and the posterior hidden state $\mathbf{h}^*$.

\begin{equation}
    \Delta \mathbf{W_{out}} = - \lambda_{out} \mathbf{\epsilon}_t \cdot \tanh(\mathbf{h}_t)^\intercal 
\end{equation}
\begin{equation}
    \begin{aligned}
        \Delta \mathbf{W_p} &= - \lambda_p \tanh(\mathbf{h}_{t-1}^*) \\
        &\cdot ((\mathbf{W_c} \cdot \mathbf{c}_{t-1})(\mathbf{W_f} \cdot \mathbf{\epsilon}_t'))^\intercal
    \end{aligned}
\end{equation}
\begin{equation}
    \begin{aligned}
            \Delta \mathbf{W_f} &= - \lambda_f \mathbf{\epsilon}_t' \\
            &\cdot ((\mathbf{W_c} \cdot \mathbf{c}_{t-1})(\mathbf{W_p} \cdot \tanh(\mathbf{h}_{t-1}^*)))^\intercal
    \end{aligned}
\end{equation}
\begin{equation}
    \begin{aligned}
        \Delta \mathbf{W_c} &= - \lambda_c \mathbf{c}_{t-1} \\
        &\cdot ((\mathbf{W_f} \cdot \mathbf{\epsilon}_t')(\mathbf{W_p} \cdot \tanh(\mathbf{h}_{t-1}^*)))^\intercal
    \end{aligned}
\end{equation}

Where $\lambda_{out}$, $\lambda_p$, $\lambda_f$, $\lambda_c$ are the learning rates for the different weight matrices of our model. The model we present here is generic, and our complete architecture features two parallel instances of this model, for the prediction of visual observations and the generation of motor commands, where the output variables are respectively denoted by $\mathbf{o}^v$ and $\mathbf{m}$ instead of $\mathbf{x}$. 

\subsection{Active inference control}
\label{sec:active_inference_controller}

As explained before, we do not have direct supervision for the generation of motor sequences. Instead, we will control at each time step the output value $\mathbf{m}_t$ of our RNN using active inference, which will provide us with a corrected motor command $\mathbf{m}_t^*$. Active inference frames motor control as a minimization of prediction error (or surprise). Here, the prediction error is the distance between the observation predicted by the visual RNN, and the observation predicted through the forward model $f$ introduced earlier, as represented in figure \ref{fig:model_overview}. To be able to infer which action might cause this error to decrease, we use the forward model that maps actions to predicted observations. The corrected motor command $\mathbf{m}_t^*$ is thus computed as a gradient descent update on $\mathbf{m}_t$ :

\begin{align}
    \mathbf{m_t^*} &= \mathbf{m_t} - \alpha_m \nabla_{\mathbf{m_t}} \|\mathbf{o_t^m} - \mathbf{o_t^v}\|_2^2 \\
    &= \mathbf{m_t} - \alpha_m \nabla_{\mathbf{m_t}} \|f(\mathbf{m_t}) - \mathbf{o_t^v}\|_2^2
\label{eq:active_inference}
\end{align}

Where $\mathbf{o_t^m}$ denotes the observation predicted from the motor pathway, and $\mathbf{o_t^v}$ denotes the observation predicted from the visual pathway.

\section{Experiments}
\label{sec:experiments}

\begin{figure*}[!ht]
    \centering
    \begin{subfigure}{0.3\textwidth}
        \centering
        \includegraphics[width=\textwidth]{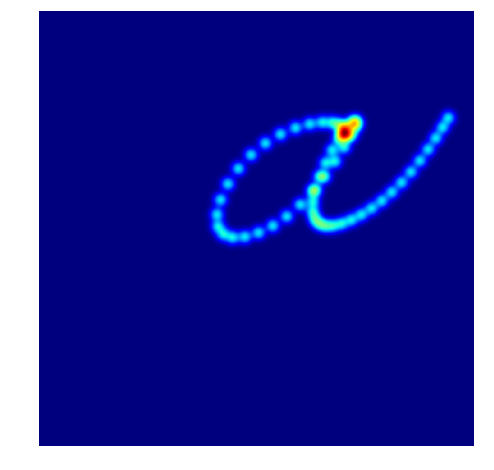}
    \end{subfigure}%
    ~
    \begin{subfigure}{0.3\textwidth}
        \centering
        \includegraphics[width=\textwidth]{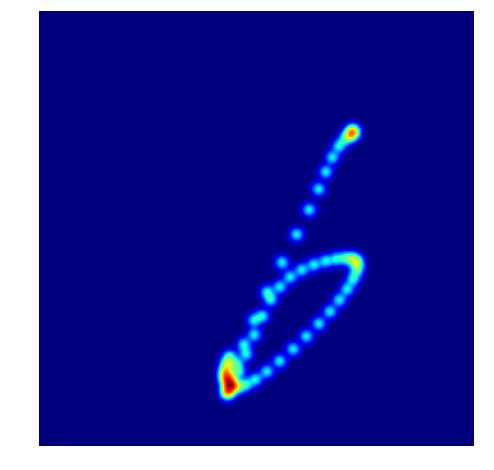}
    \end{subfigure}%
    ~
    \begin{subfigure}{0.3\textwidth}
        \centering
        \includegraphics[width=\textwidth]{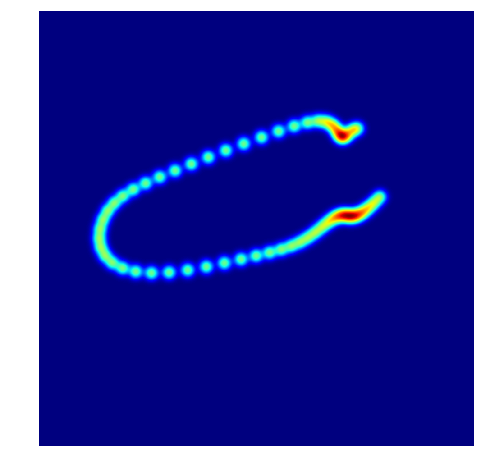}
    \end{subfigure}%
    \caption{Predicted sequences of 2D positions at the end of training for the three given classes \textit{a} (left), \textit{b} (middle) and \textit{c} (right). Each predicted position is represented as a Gaussian centered on it. The heatmaps represent the sum of these Gaussians for each trajectory.}
    \label{fig:heatmap}
\end{figure*}

In this section, we perform several experiments to validate our approach. We start by testing the RNN model presented in section \ref{sec:predictive_coding_rnn} on the task of modeling 2D handwriting trajectories, and highlight the features of this model that could be interesting for motor control. Second, we validate the whole model by demonstrating that our architecture is able to learn to generate robust motor trajectories for handwriting. We compare our model's encoding capacity to the algorithm presented in \cite{Mochizuki2013}. We experiment with different situations that highlight the interest of the online interaction of the two modalities modeled in our architecture.

\subsection{Predictive coding RNN}

In this subsection, we experiment with isolated instances of the RNN model presented in section \ref{sec:predictive_coding_rnn}. We first explore the properties of this model used for the prediction and inference of visual trajectories. Then, we investigate the relevance of this model for the generation of motor trajectories.

\subsubsection{Visual prediction of handwritten trajectories}
\label{sec:rnn_experiment}

We first report the results obtained for the prediction of trajectories corresponding to handwritten letters. The model is trained using 2D trajectories from the Character Trajectories Data Set of the UCI Machine Learning Repository \cite{Dua2019} as target observations $\mathbf{o}^*$. We use up to 16 classes from the data set, each class corresponding to a different letter of the alphabet. We take 20 trajectories from each class to form the training set, and 20 other trajectories to form the testing set. All trajectories are rescaled to last 60 time steps.

We used for this experiment a hidden state $\mathbf{h}^v$ of dimension 50, and hidden causes $\mathbf{c}^v$ of dimension $p$, where $p$ is the number of trajectory classes to learn (between 1 and 16 depending on the experiment). Intuitively, we would expect the hidden causes variable to correspond to a certain representation of the predicted trajectories. In our supervised learning setup, we enforced the hidden causes value for each trajectory as the one-hot vector corresponding to the trajectory label (i.e. a vector of dimension $p$ filled with 0 except for a 1 on the index corresponding to the trajectory label). This was made possible by initializing the hidden causes to this value at the beginning of each trajectory, and cancelling the hidden causes update by setting the value of $\alpha_c$ in equation \ref{eq:hidden_causes_update} to 0. The initial hidden state is sampled randomly from a normal distribution and is shared for all trajectories. We trained our RNN on three letter categories, corresponding to the labels \textit{a}, \textit{b}, and \textit{c}. When used for \textit{prediction}, the backward propagation of prediction error is cancelled out by setting to 0 the values of $\alpha_h$ and $\alpha_c$. The predicted trajectory is completely conditioned by the initial hidden state and hidden causes. Figure \ref{fig:heatmap} displays the predicted trajectories as heatmaps, for each possible initial hidden causes \textit{a}, \textit{b}, or \textit{c}. After training, the RNN is able to properly predict the trajectory without any supervision.  

Contrary to classical RNN models, the proposed architecture can perform \textit{inference} as well as \textit{prediction}. When used for \textit{inference}, the hidden states and hidden causes are updated by the backward propagation of prediction error based on given visual observations $\mathbf{o}^*$. After each update, we enforce the hidden causes to encode a multinomial probability distribution on the trajectory classes by clipping the values between $[0, 1]$ and normalizing the vector to sum to 1. The initial hidden causes are chosen to encode a uniform discrete distribution over the possible classes by setting their values to $\frac{1}{p}$.

\begin{figure*}[!ht]
    \centering
    \begin{subfigure}{0.24\textwidth}
        \centering
        \includegraphics[width=\textwidth]{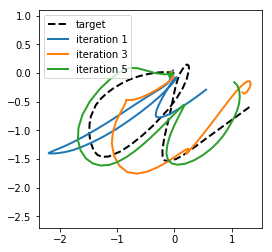}
        \caption{Target trajectory provided (black) and trajectories predicted by the RNN after 1, 3 and 5 presentations of the target.}
        \label{fig:inference_a_traj}
    \end{subfigure}%
    ~
    \begin{subfigure}{0.23\textwidth}
        \centering
        \includegraphics[width=\textwidth]{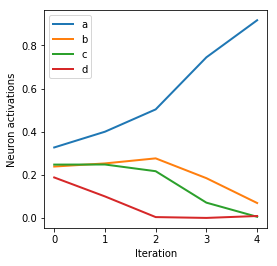}
        \caption{Evolution of the hidden causes neuron activations for the 4 categories, according to the number of presentations of the target trajectory.}
        \label{fig:inference_a_act}
    \end{subfigure}%
    ~
    \begin{subfigure}{0.24\textwidth}
        \centering
        \includegraphics[width=\textwidth]{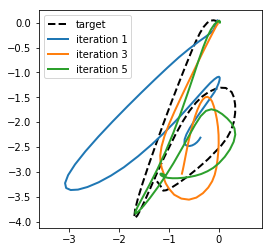}
        \caption{Target trajectory provided (black) and trajectories predicted by the RNN after 1, 3 and 5 presentations of the target.}
        \label{fig:inference_b_traj}
    \end{subfigure}%
    ~
    \begin{subfigure}{0.23\textwidth}
        \centering
        \includegraphics[width=\textwidth]{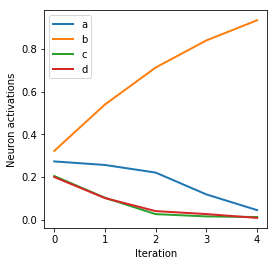}
        \caption{Evolution of the hidden causes neuron activations for the 4 categories, according to the number of presentations of the target trajectory.}
        \label{fig:inference_b_act}
    \end{subfigure}%
    \caption{Inference of trajectory labels given targets. The inference process is demonstrated with the evolution of the predicted trajectories (\textbf{a}, \textbf{c}) and the evolution of the hidden causes (\textbf{b}, \textbf{d}) during 5 presentation of the target trajectory.}
    \label{fig:inference}
\end{figure*}

\begin{figure*}[!ht]
    \centering
    \begin{subfigure}{0.3\textwidth}
        \centering
        \includegraphics[width=\textwidth]{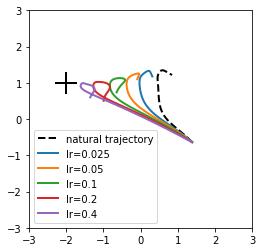}
        \caption{Controlled RNN trajectories, labels correspond to the learning rate value $\alpha_h$ used in equation \ref{eq:hidden_state_update}, weighting the influence on the bottom-up signal on state update. The dashed line represents the natural RNN trajectory.}
        \label{fig:exp1_controlled_trajectories}
    \end{subfigure}%
    ~
    \begin{subfigure}{0.3\textwidth}
        \centering
        \includegraphics[width=\textwidth]{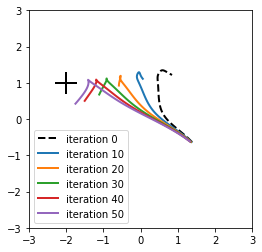}
        \caption{Evolution of the natural RNN trajectories throughout learning, labels correspond to training iterations. The dashed line represents the natural RNN trajectory before training. These trajectories were obtained with $\alpha_h=0.025$.}
        \label{fig:exp1_learning}
    \end{subfigure}%
    ~
    \begin{subfigure}{0.3\textwidth}
        \centering
        \includegraphics[width=\textwidth]{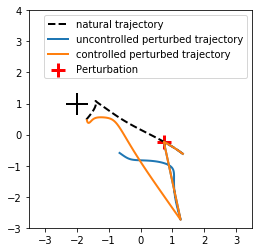}
        \caption{Effect of a perturbation onto the trajectory, after training, with or without control. The perturbation is applied at time step 20, and represented as a red segment. The controlled trajectory is obtained with $\alpha_h = 0.2$.}
        \label{fig:exp1_perturbation}
    \end{subfigure}%
    \caption{Results for motor control in the simplified setup. The "plus" shaped marker represents the target position used by the active inference controller. All trajectories start with the same RNN hidden state and hidden causes.}
    \label{fig:exp1}
\end{figure*}

Figure \ref{fig:inference} displays the process of \textit{inference} of the category of a given trajectory, after training. The target trajectories belong to the testing set, and represent the letters \textit{a} and \textit{b}. By repetitively presenting the target visual trajectory to the network, the hidden causes variable converges towards the one-hot vector corresponding to the proper class. This experiment shows that our RNN model can learn to predict visual trajectories given associated labels, but also infer labels given the associated visual trajectories. It can thus be used for both generation and classification.

\subsubsection{Sandbox experiment for motor control}
\label{sec:sandbox}

\begin{figure*}[!ht]
    \centering
    \begin{subfigure}{0.31\textwidth}
        \centering
        \includegraphics[width=\textwidth]{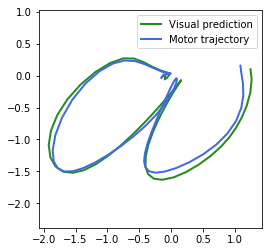}
    \end{subfigure}%
    ~
    \begin{subfigure}{0.3\textwidth}
        \centering
        \includegraphics[width=\textwidth]{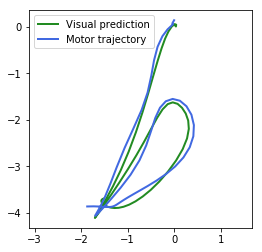}
    \end{subfigure}%
    ~
    \begin{subfigure}{0.3\textwidth}
        \centering
        \includegraphics[width=\textwidth]{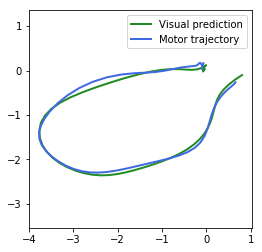}
    \end{subfigure}%
    \caption{Generated motor trajectories and predicted visual sequences of 2D positions at the end of training for the three given classes \textit{a} (left), \textit{b} (middle) and \textit{c} (right). Results obtained with $n=50$ and $p=3$.}
    \label{fig:motor_prediction_examples}
\end{figure*}

Now that we have demonstrated the relevance of our model for the visual prediction and perceptual inference of handwritten trajectories, we investigate its relevance for motor control.

In this sandbox experiment, we highlight our model's ability to perform online adaptation given a target. We start with a simplified situation, temporarily ignoring the constraints put forward in the introduction, and suppose that direct supervision in the "motor" space is available. For visualization purposes, we suppose that the motor space is a 2D euclidean space, and that the target is constant.

First, the motor RNN weights are initialized randomly. The RNN exhibits natural (i.e. uncontrolled, without feedback) trajectories as represented in dashed lines in figure \ref{fig:exp1_controlled_trajectories} and \ref{fig:exp1_learning}. The target motor command $\mathbf{m}^*$ is represented by the "plus" shaped marker. Figure \ref{fig:exp1_controlled_trajectories} shows that the bottom-up prediction error signal is able to adjust the trajectories towards the target, modulated by the coefficient $\alpha_h$ introduced in equation \ref{eq:hidden_state_update}. This online adaptation is an interesting feature for motor generative models, as it could allow it to resist to noise or perturbations. In comparison to the previous experiment where online update of hidden states of the generative model was seen as a perceptual inference process, we draw here a parallel with motor control.

Figure \ref{fig:exp1_learning} displays the evolution of the natural trajectories the RNN exhibits throughout learning. At the end of learning, the natural RNN trajectory seems to be attracted towards the neighbourhood of the target position. Figure \ref{fig:exp1_perturbation} shows the effect of a perturbation applied onto the RNN hidden state during the generation of a trajectory, after learning. In the controlled case, the trajectory still converges towards the target position.

These first results show that our RNN generative model can learn and control simple trajectories in a 2D space.

\subsection{Experiments on the complete architecture}
\label{sec:architecture_experiments}

In this subsection, we experiment with the complete architecture presented in figure \ref{fig:model_overview}. This architecture is composed of two RNN models for the generation of the motor commands and visual predictions, and a forward model $f$ translating motor commands $\mathbf{m}$ into expected resulting visual observations $\mathbf{o}^m$.

Figure \ref{fig:brain_arm} represents our experimental setup for this experiment. Our agent evolves in an environment with which it can interact through sensors and actuators. Since we focus here on motor skill learning, we simplified the visual space of our agent by already decoding the position of the agent's end-effector from its visual input. In other words, the agent directly receives as visual input the position of its end-effector in Cartesian coordinates. The agent acts on the 2D environment by moving the 3 degrees of freedom simulated arm. The extremity of the first joint (the shoulder) is fixed on the position (-6, 6). The three joints are of respective lengths 6, 4 and 2. We don't cover here the learning of the forward model, and suppose that the agent has learned through motor babbling how its actions influence its observations. In our experiments, the forward model is replaced by the real physical model outputting end-effector positions in Cartesian coordinates according to joint orientations. Our architecture's training is thus composed of two stages:

\begin{enumerate}
    \item Supervised learning of visual handwritten trajectories : The RNN predicting trajectories in the visual space (2 dimensions) is trained as already showed in section \ref{sec:rnn_experiment}.
    \item Training of the motor RNN : The RNN generating trajectories in the motor space (3 dimensions) is trained using the method described in section \ref{sec:active_inference_controller}, using the trajectories predicted by the visual RNN as indirect supervision to perform AIF.
\end{enumerate}

\subsubsection{Motor generation of handwritten trajectories}

We now focus on the training of the motor RNN, using the learned visual prediction RNN. The visual prediction RNN generates a trajectory of observations that the motor RNN tries to replicate. We train the motor RNN according to the method detailed in section \ref{sec:active_inference_controller}. The motor RNN hidden state and hidden causes are initialized following the same procedure as the visual prediction RNN: hidden states are initialized randomly and shared across the different classes of trajectories. The initial hidden causes are one-hot vectors of dimension $p$ (where $p$ is the number of classes) encoding the trajectory label. Figure \ref{fig:motor_prediction_examples} displays learned motor trajectories along with the corresponding predicted visual trajectories. These results were obtained with a hidden state dimension of $n=50$ for both RNNs and with $p=3$ classes, after 40000 iterations on the training set. Training is arguably long with regard to the difficulty of the proposed task. We suppose that this is due to the PC learning mechanism, that can only approximate backpropagation properly if we let enough time for the inference of each variable in the computation graph to converge \cite{Millidge2020}.

\subsubsection{Model capacity}

\begin{figure}[!ht]
\centering
    \includegraphics[width=\columnwidth]{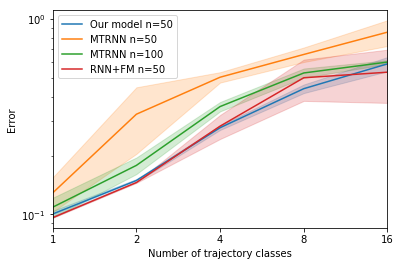}
    \caption{Comparison of the reconstruction error according to the number of trajectory classes for three models : our model with $n=50$ and two instances of the MTRNN model proposed in \cite{Mochizuki2013} with $n=50$ and $n=100$. The curves display the average reconstruction error on the testing data set. Areas in transparency indicate confidence intervals.}
    \label{fig:motor_capacity}
\end{figure}

Before analyzing other behaviours of the proposed model, we validate it by comparing its performance with two benchmark models. 

First, we compare our performance with the method presented in \cite{Mochizuki2013}, proposed on a similar task. This method uses a Multiple Timescales RNN (MTRNN) \cite{Yamashita2008} model for the joint generation of visual and motor trajectories as represented in figure \ref{fig:habituated_trajectories_model}. The model is first trained using visuomotor trajectories obtained through motor babbling. Then, optimization of the initial MTRNN hidden state is performed for each target visual trajectory using BPTT. Finally, the MTRNN weights are tuned to ensure coherence between the visual and motor outputs. The detailed algorithm is presented in appendix \ref{annex:mtrnn_algo}.

Second, we compare our model with a simple architecture composed of an RNN generating motor trajectories and a forward model (equivalent to the one present in our model) translating this motor output into visual trajectories. Indeed, we could argue that using the forward model we introduced, one could simply backpropagate gradients originating from an error signal in the visual space to learn motor trajectories. We train such a model, that we label RNN+FM, using BPTT, and compare its performance with our model.

Performances of the three models are evaluated according to their precision on the motor trajectory reconstruction, and their capacity to encode a large number of trajectories. Precision is measured with the average reconstruction error on the testing data set. Figure \ref{fig:motor_capacity} displays the evolution of this error measure for our model, two instances of the MTRNN model, and the RNN+FM model of the comparison model.

If we extract the motor RNN model from the architecture, and compare it with a MTRNN model with the same hidden state dimension, both models have a comparable number of parameters. However, we could argue that the same MTRNN can generate trajectories in both visual and motor space, while our motor RNN only generates trajectories in the motor space. For fair comparison, we might want to include into the parameter count the parameters of the visual RNN in our architecture. For this reason, we extend the comparison with a MTRNN model with a state dimension of $n=100$, with twice as many parameters as our two RNNs combined. Finally, note that our approach, contrary to the MTRNN model, assumes that a perfect forward model is available to perform AIF.

Still, the results displayed in figure \ref{fig:motor_capacity} tend to show that our architecture can compete with other algorithms for motor trajectory learning with indirect supervision.

\subsubsection{Intermittent control}

\begin{figure*}[!ht]
    \centering
    \includegraphics[width=0.9\textwidth]{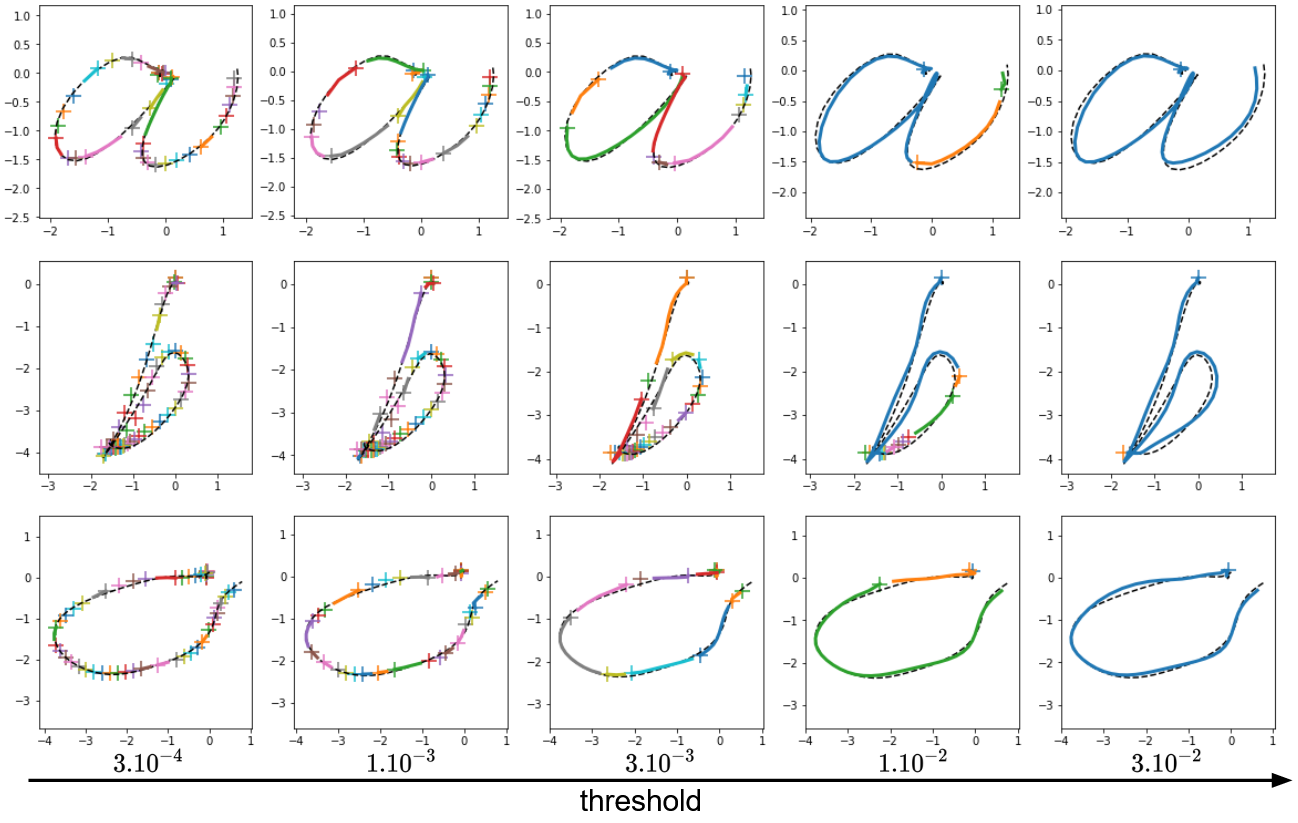}
    \caption{Trajectories generated by the motor RNN, with a state dimension of 50, displayed into the visual space. The black dashed line represents the visual trajectory predicted by the visual RNN. The trajectories generated by the motor RNN are represented as successions of trajectory segments, differentiated by their color. Each new trajectory segment corresponds to an activation of the active inference controller, and is represented by a plus shaped marker. The different figures correspond to different activation threshold values for the controller, from left to right: $3.10^{-4}, 1.10^{-3}, 3.10^{-3}, 1.10^{-2}, 3.10^{-2}$.}
    \label{fig:intermittent_control}
\end{figure*}

One of our model's feature not discussed previously is the possibility to switch off the feedback pathway when prediction error is under a certain threshold. We experimented with this idea by varying such a threshold and observing when the feedback pathway would switch on and off. Figure \ref{fig:intermittent_control} displays results that were obtained with a hidden state dimension of $n=50$ with $p=3$ classes, after training of the motor RNN.

For the lowest threshold value, the feedback pathway is almost always active and the motor trajectories are controlled to accurately match the predicted visual trajectories. For the highest threshold value, the feedback pathway never activates and the trajectory performed correspond to the natural trajectory of the motor RNN. This mechanism is interesting as it could be used to control the trade-off between precision and smoothness of the generated trajectories in situations where the visual target trajectory isn't smooth.  

\subsubsection{Perturbation robustness}

\begin{figure*}[!ht]
    \centering
    \begin{subfigure}{0.285\textwidth}
        \centering
        \includegraphics[width=\textwidth]{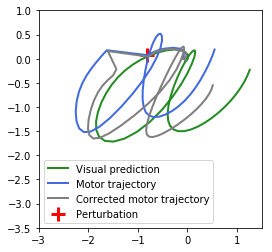}
        \caption{Motor trajectory corresponding to the letter \textit{a} with and without control for $\sigma_p^2=0.6$.}
    \end{subfigure}%
    ~
    \begin{subfigure}{0.275\textwidth}
        \centering
        \includegraphics[width=\textwidth]{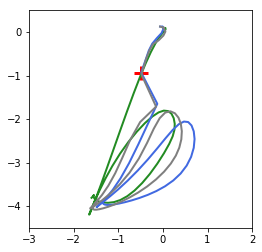}
        \caption{Motor trajectory corresponding to the letter \textit{b} with and without control for $\sigma_p^2=0.6$.}
    \end{subfigure}%
    ~
    \begin{subfigure}{0.44\textwidth}
        \centering
        \includegraphics[width=\textwidth]{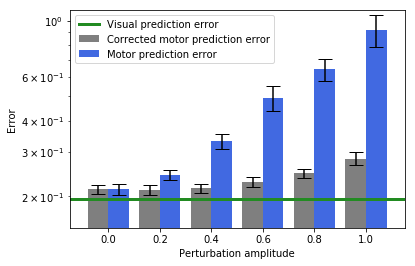}
        \caption{Average prediction error according to the perturbation amplitude $\sigma_p^2$. Prediction errors are measured as average distances with regard to corresponding test trajectories of the same label.}
    \end{subfigure}%
    \caption{Perturbation robustness experiment. At the 10th time step, we apply a random perturbation, represented in red, on the motor output. Our model uses the visual prediction as a guide to correct the perturbed motor trajectory.}
    \label{fig:perturbation_experiment}
\end{figure*}

\begin{figure*}[!ht]
    \centering
    \begin{subfigure}{0.285\textwidth}
        \centering
        \includegraphics[width=\textwidth]{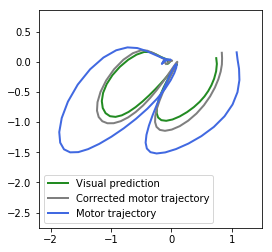}
        \caption{Motor trajectory corresponding to the letter \textit{a} with and without control for a scaling factor of $0.6$.}
    \end{subfigure}%
    ~
    \begin{subfigure}{0.275\textwidth}
        \centering
        \includegraphics[width=\textwidth]{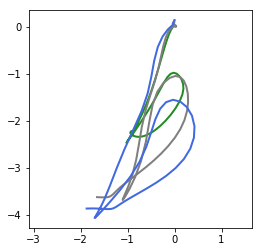}
        \caption{Motor trajectory corresponding to the letter \textit{b} with and without control for a scaling factor of $0.6$.}
    \end{subfigure}%
    ~
    \begin{subfigure}{0.42\textwidth}
        \centering
        \includegraphics[width=\textwidth]{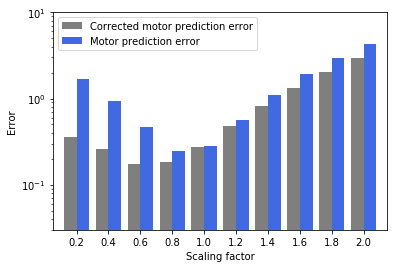}
        \caption{Average prediction error according to the scaling factor applied on the visual prediction. Prediction errors are measured as average distances with regard to corresponding rescaled test trajectories of the same label.}
        \label{fig:scaling_results}
    \end{subfigure}%
    \caption{Adaptation to scaling experiment. We apply a change of scale on the visual prediction. Our model dynamically adapts the motor trajectory to fulfill best the visual prediction.}
    \label{fig:scaling_experiment}
\end{figure*}

\begin{figure*}[!ht]
    \centering
    \begin{subfigure}{0.285\textwidth}
        \centering
        \includegraphics[width=\textwidth]{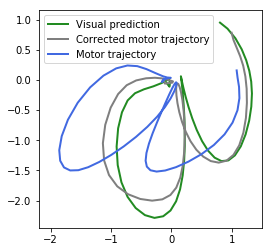}
        \caption{Motor trajectory corresponding to the letter \textit{a} with and without control for a rotation of angle $\pi/4$.}
    \end{subfigure}%
    ~
    \begin{subfigure}{0.275\textwidth}
        \centering
        \includegraphics[width=\textwidth]{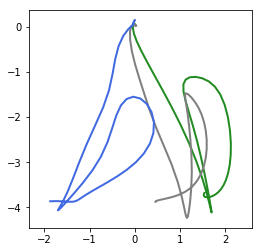}
        \caption{Motor trajectory corresponding to the letter \textit{b} with and without control for a rotation of angle $\pi/4$.}
    \end{subfigure}%
    ~
    \begin{subfigure}{0.42\textwidth}
        \centering
        \includegraphics[width=\textwidth]{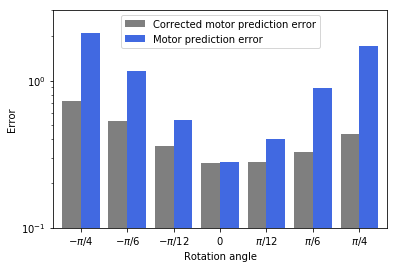}
        \caption{Average prediction error according to the angle of the rotation applied to the visual prediction. Prediction errors are measured as average distances with regard to corresponding rotated test trajectories of the same label.}        \label{fig:rotation_results}
    \end{subfigure}%
    \caption{Adaptation to rotation experiment. We apply a rotation on the visual prediction. Our model dynamically adapts the motor trajectory to fulfill best the visual prediction.}
    \label{fig:rotation_experiment}
\end{figure*}

As previously showed in section \ref{sec:sandbox}, the motor RNN should be able to dynamically adapt to perturbations. In this experiment, we consider applying external perturbations of variable amplitude onto the motor output of our model. Perturbations sampled from a multivariate normal distribution of variance $\sigma_p^2 \mathbf{I_3}$ are added to the motor output of the generative model for all timesteps $t>10$. Figure \ref{fig:perturbation_experiment} displays examples of obtained trajectories with or without control, and the evolution of the average prediction error with regard to the perturbation amplitude $\sigma_p^2$.

These results demonstrate that our model generates motor trajectories robust to external perturbations.

\subsubsection{Adaptation to rescaling and rotation}

\begin{figure*}[!ht]
    \centering
    \begin{subfigure}{0.295\textwidth}
        \centering
        \includegraphics[width=\textwidth]{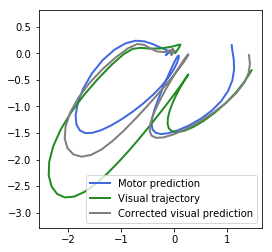}
        \caption{Predicted visual trajectory corresponding to the letter \textit{a} with and without correction for $\sigma_i^2=0.1$.}
    \end{subfigure}%
    ~
    \begin{subfigure}{0.285\textwidth}
        \centering
        \includegraphics[width=\textwidth]{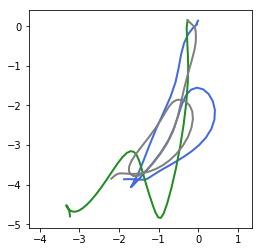}
        \caption{Predicted visual trajectory corresponding to the letter \textit{b} with and without correction for $\sigma_i^2=0.1$.}
    \end{subfigure}%
    ~
    \begin{subfigure}{0.42\textwidth}
        \centering
        \includegraphics[width=\textwidth]{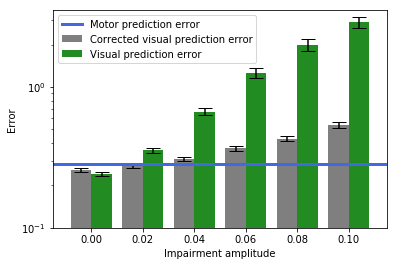}
        \caption{Average prediction error according to the impairment amplitude $\sigma_i^2$. Prediction errors are measured as average distances with regard to corresponding test trajectories of the same label.}
    \end{subfigure}%
    \caption{Impairments experiment. The visual RNN is impaired by applying a multiplicative noise of varying amplitude onto its parameters. Our model uses the motor prediction as a guide to correct the visual trajectory.}
    \label{fig:forgetting_experiment}
\end{figure*}

In this experiment, we consider the situation where a transformation is applied on the predicted visual trajectory. Changes of scales and rotations are transformations that can be applied easily on the visual output. However, generating the motor commands performing these transformed visual trajectories is less trivial. We experiment here applying changes of scales and orientations to the visual prediction while controlling the motor trajectory using the prediction error feedback. Results of those experiments for different scales and orientations are displayed respectively in figures \ref{fig:scaling_experiment} and \ref{fig:rotation_experiment}.

Figure \ref{fig:scaling_results} displays the evolution of the error between the motor trajectories and the rescaled test trajectories, with and without control. We first notice that overall, reducing the scale seems to induce lesser prediction errors than increasing the scale. This is because the prediction error is computed as a path integral of point to point distances, thus sensible to the scale of the trajectories. Second, we notice that the improvement brought by the online control is less effective when increasing the scale. This is due to the fact that the motor RNN was trained to exhibit trajectories of limited amplitudes. Inference of the hidden state of the RNN is not sufficient to properly control the trajectory.

Figure \ref{fig:rotation_results} displays the evolution of the error between the motor trajectories and the rotated test trajectories, with and without control. We notice that the architecture manages to adapt better to rotations in the counterclockwise direction. This could be due to the fact that performing these trajectories requires smaller modifications on the natural motor trajectory because of the arm configuration.

\subsubsection{Impairments of the visual prediction RNN}

\begin{figure}[!ht]
    \centering
    \begin{subfigure}{0.22\textwidth}
        \centering
        \includegraphics[width=\textwidth]{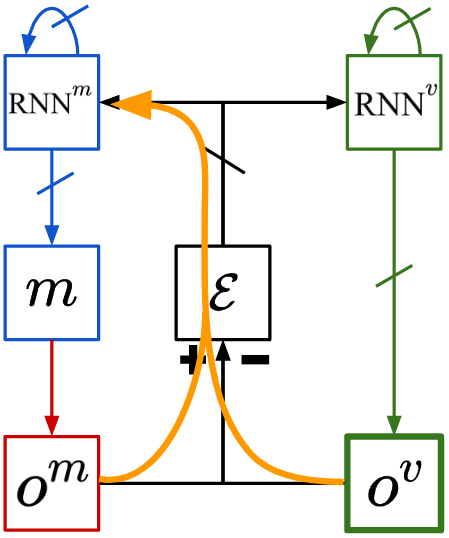}
        \caption{Visual to motor feedback pathway. Generated motor trajectories are controlled based on the visual prediction.}
        \label{fig:visual_to_motor}
    \end{subfigure}%
    ~
    \begin{subfigure}{0.22\textwidth}
        \centering
        \includegraphics[width=\textwidth]{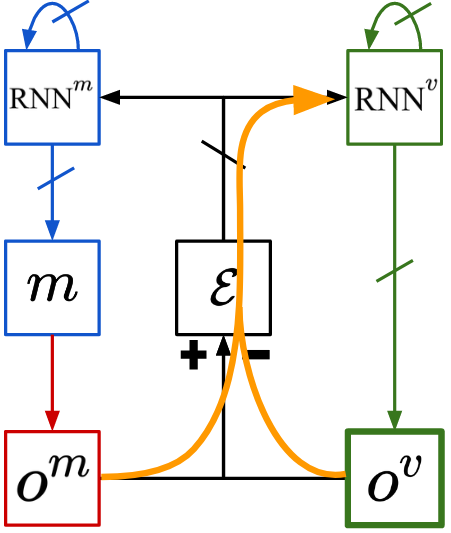}
        \caption{Motor to visual feedback pathway. Predicted visual trajectories are corrected based on the motor prediction.}
        \label{fig:motor_to_visual}
    \end{subfigure}%
    \caption{Prediction error feedback pathways.}
    \label{fig:feedback_pathways}
\end{figure}

In all the previous experiments of this section, we have considered only one feedback pathway between the two possible feedback loops in our architecture, represented in figure \ref{fig:feedback_pathways}. Whether it was to learn motor trajectories, for intermittent control, to adapt to perturbations on the motor output, or to adapt to transformations on the visual prediction, we only used the visual to motor feedback pathways (figure \ref{fig:visual_to_motor}). Since the motor RNN was trained using the visual prediction RNN, using the motor predictions $\mathbf{o}^m$ to control for the visual predictions $\mathbf{o}^v$ would only result in less precise visual predictions.

However, the symmetry of the interaction between the two modalities can still be exploited in situations where the visual prediction RNN performs worse than the motor RNN. To create such situations, we impaired the visual prediction RNN by applying a multiplicative noise $\mathcal{N}(1, \sigma_i^2)$ to the model's parameters $\mathbf{W_{out}}$, $\mathbf{W_p}$, $\mathbf{W_f}$, $\mathbf{W_c}$. We argue that this situation can arise naturally if we consider the lifelong learning of an agent. The impairment we simulated here could correspond to the forgetting of the visual prediction RNN, that could be due to training on a different task.

Similarly to the previous experiments, we display in figure \ref{fig:forgetting_experiment} examples of predicted visual trajectories with or without correction from the motor RNN, and the evolution of the average prediction error with regard to the impairment amplitude $\sigma_i^2$. These results demonstrate that knowledge in our architecture can be transferred in both directions.

We have only considered the collaborative situation where the two RNN models had the same prior belief about the trajectory being written, i.e. the same initial hidden causes in our architecture. We could also conceive the situation where both systems compete to convince the other of its own belief. In this hypothesised situation, not experimented in this article, both feedback pathways would be used at the same time.

\section{Discussion}

We have shown how an architecture can learn motor trajectories from an indirect supervision in the visual space. The adaptive behaviour resulting from the addition of prediction error bottom-up pathways, and the bidirectional interactions between the vision prediction RNN and the motor RNN, give our model the ability to dynamically control motor trajectories according to visual predictions. We have evaluated our model in different experimental setups highlighting interesting features such as intermittent control, robustness to perturbation, and adaptation to scaling and rotation.

These properties are reminiscent of the behaviors sought by the line of research on Dynamical Movement Primitives (DMP) \cite{Ijspeert2013}, aiming at modeling attractor behaviors for autonomous nonlinear dynamical systems. The essence of the approach is to define a simple dynamical system, and tune it so that it exhibits prescribed attractor dynamics by means of a learnable forcing term. The online control performed by the prediction error minimization process in our model gives rise to similar attractor behaviors when confronted with external perturbations.

The clear separation between the sensory path and the motor path could allow our agent to learn motor trajectories with different forward models (for instance right hand and left hand), based on a unique representation of the trajectory in the visual space. This can also be useful in case of modifications of the robot's limbs, for instance by removing (amputation) or adding a joint (use of a tool).

The bidirectionality of the proposed architecture can also allow a knowledge transfer from the motor system to the visual system. A notable limitation is that we had to manually set the different learning rate coefficients $\alpha_h^v$, $\alpha_h^m$, $\alpha_c^v$ and $\alpha_c^h$ to direct the influence of the two networks onto each other. A possible solution could be the estimation of the variance of each random variable in the generative models to account for different levels of certainty of priors. The influence between the two networks would then naturally be directed from the system with the highest certainty to the one with the lowest. This property, called precision weighting, can naturally result from the modelisation of hidden states as random variables distribution as proposed in the free energy formulation of PC \cite{Friston2009c}.

Another important limitation of this work is the assumption that a perfect forward model is available. The natural solution to this limitation would be to include the forward model learning into our architecture. Interestingly, the forward model could also incorporate hidden causes variables that could be inferred dynamically, allowing for fast adaptations to modifications of the forward model such as limb amputation or tool use. Adaptation to variability of body layouts have been investigated previously in \cite{Abrossimoff2018, Braud2018}.

The proposed model learns motor trajectories according to corresponding trajectories in its sensory space. In future work, we could imagine using the same approach to learn motor trajectories in more complex environments, for instance manipulation of objects with a robotic arm using visual feedback, or navigation in an environment with first-person view.

Future work could also investigate more deeply into the behavior of the RNN model proposed in section \ref{sec:predictive_coding_rnn}. We have shown that the modelisation of hidden causes gives our model the ability to dynamically perform inference of a representation of the provided trajectories, making it a system capable of performing both generation and classification of sequences. In particular, the fact that different hidden causes generate different hidden state dynamics could allow this model to generate a large variety of target trajectories, compared to standard RNNs with fixed dynamics. Preliminary work in this direction tends to show that this model could exhibit interesting dynamics known as chaotic itinerancy \cite{Tsuda2015, Inoue2020}.

\bibliographystyle{ieeetr}
\bibliography{references.bib}

\begin{thebibliography}{10}

\bibitem{Hochreiter1997}
S.~Hochreiter and J.~Schmidhuber, ``Long short-term memory,'' {\em Neural
  Computation}, vol.~9, no.~8, pp.~1735--1780, 1997.

\bibitem{Sutskever2009}
I.~Sutskever, G.~E. Hinton, and G.~W. Taylor, ``The recurrent temporal
  restricted boltzmann machine,'' in {\em Advances in Neural Information
  Processing Systems}, vol.~21, pp.~1601--1608, 2009.

\bibitem{Williams1989}
R.~J. Williams and D.~Zipser, ``Experimental analysis of the real-time
  recurrent learning algorithm,'' {\em Connection Science}, vol.~1, no.~1,
  pp.~87--111, 1989.

\bibitem{Verstraeten2007}
D.~Verstraeten, B.~Schrauwen, M.~D’Haene, and D.~Stroobandt, ``{An
  experimental unification of reservoir computing methods},'' {\em Neural
  Network}, vol.~20, pp.~391--403, 2007.

\bibitem{Lukosevicius2009}
M.~Lukoševičius and H.~Jaeger, ``Reservoir computing approaches to recurrent
  neural network training,'' {\em Computer Science Review}, vol.~3, no.~3,
  pp.~127 -- 149, 2009.

\bibitem{Friston2009a}
K.~Friston, J.~Daunizeau, and S.~Kiebel, ``Reinforcement learning or active
  inference?,'' {\em PLoS ONE}, vol.~4, no.~7, p.~e6421, 2009.

\bibitem{Friston2016}
K.~Friston, T.~FitzGerald, F.~Rigoli, P.~Schwartenbeck, J.~O’Doherty, and
  G.~Pezzulo, ``Active inference and learning,'' {\em Neuroscience \&
  Biobehavioral Reviews}, vol.~68, pp.~862--879, 2016.

\bibitem{Friston2006}
K.~Friston and J.~Kilner, ``A free energy principle for the brain,'' {\em J.
  Physiol. Paris}, vol.~100, pp.~70--87, 2006.

\bibitem{Buckley2017}
C.~L. Buckley, C.~S. Kim, S.~McGregor, and A.~K. Seth, ``The free energy
  principle for action and perception: A mathematical review,'' {\em Journal of
  Mathematical Psychology}, vol.~81, pp.~55 -- 79, 2017.

\bibitem{Mochizuki2013}
K.~{Mochizuki}, S.~{Nishide}, H.~G. {Okuno}, and T.~{Ogata}, ``Developmental
  human-robot imitation learning of drawing with a neuro dynamical system,'' in
  {\em 2013 IEEE International Conference on Systems, Man, and Cybernetics},
  pp.~2336--2341, 2013.

\bibitem{Sasaki2019}
K.~{Sasaki} and T.~{Ogata}, ``Adaptive drawing behavior by visuomotor learning
  using recurrent neural networks,'' {\em IEEE Transactions on Cognitive and
  Developmental Systems}, vol.~11, no.~1, pp.~119--128, 2019.

\bibitem{Rao1999}
R.~Rao and D.~Ballard, ``Predictive coding in the visual cortex a functional
  interpretation of some extra-classical receptive-field effects,'' {\em Nat
  Neurosci}, vol.~2, pp.~79--87, 1999.

\bibitem{Clark2013}
A.~Clark, ``Whatever next? predictive brains, situated agents, and the future
  of cognitive science,'' {\em Behavioral and Brain Sciences}, vol.~36, no.~3,
  p.~181–204, 2013.

\bibitem{Ororbia2020}
A.~{Ororbia}, A.~{Mali}, C.~L. {Giles}, and D.~{Kifer}, ``Continual learning of
  recurrent neural networks by locally aligning distributed representations,''
  {\em IEEE Transactions on Neural Networks and Learning Systems}, vol.~31,
  no.~10, pp.~4267--4278, 2020.

\bibitem{Friston2009c}
K.~Friston and S.~Kiebel, ``Predictive coding under the free-energy
  principle,'' {\em Philosophical Transactions of the Royal Society of London.
  Series B, Biological Sciences}, vol.~364, pp.~1211--21, 2009.

\bibitem{Wang2010}
W.~Wang, S.~S. Chan, D.~A. Heldman, and D.~W. Moran, ``Motor cortical
  representation of hand translation and rotation during reaching,'' {\em
  Journal of Neuroscience}, vol.~30, no.~3, pp.~958--962, 2010.

\bibitem{Hoerzer2012}
G.~M. Hoerzer, R.~Legenstein, and W.~Maass, ``{Emergence of Complex
  Computational Structures From Chaotic Neural Networks Through
  Reward-Modulated Hebbian Learning},'' {\em Cerebral Cortex}, vol.~24,
  pp.~677--690, 11 2012.

\bibitem{Mannella2015}
F.~Mannella and G.~Baldassarre, ``Selection of cortical dynamics for motor
  behaviour by the basal ganglia,'' {\em Biol. Cybern.}, vol.~109,
  p.~575–595, Dec. 2015.

\bibitem{Shadmehr2010}
R.~Shadmehr, M.~A. Smith, and J.~W. Krakauer, ``Error correction, sensory
  prediction, and adaptation in motor control,'' {\em Annual Review of
  Neuroscience}, vol.~33, no.~1, pp.~89--108, 2010.
\newblock PMID: 20367317.

\bibitem{Creemregehr2009}
S.~H. Creem-Regehr, ``Sensory-motor and cognitive functions of the human
  posterior parietal cortex involved in manual actions,'' {\em Neurobiology of
  Learning and Memory}, vol.~91, no.~2, pp.~166 -- 171, 2009.
\newblock Special Issue: Parietal Cortex.

\bibitem{Planton2017}
S.~Planton, M.~Longcamp, P.~Péran, J.-F. Démonet, and M.~Jucla, ``How
  specialized are writing-specific brain regions? an fmri study of writing,
  drawing and oral spelling,'' {\em Cortex}, vol.~88, pp.~66 -- 80, 2017.

\bibitem{Pezzulo2016}
G.~Pezzulo and P.~Cisek, ``Navigating the affordance landscape: Feedback
  control as a process model of behavior and cognition,'' {\em Trends in
  Cognitive Sciences}, vol.~20, no.~6, pp.~414 -- 424, 2016.

\bibitem{Mushiake2006}
H.~Mushiake, N.~Saito, K.~Sakamoto, Y.~Itoyama, and J.~Tanji, ``Activity in the
  lateral prefrontal cortex reflects multiple steps of future events in action
  plans,'' {\em Neuron}, vol.~50, no.~4, pp.~631 -- 641, 2006.

\bibitem{Botvinick2009}
M.~Botvinick and J.~An, ``Goal-directed decision making in prefrontal cortex: A
  computational framework,'' {\em Advances in neural information processing
  systems}, vol.~21, p.~169—176, 2009.

\bibitem{Pascanu2012}
R.~Pascanu, T.~Mikolov, and Y.~Bengio, ``Understanding the exploding gradient
  problem,'' {\em CoRR}, vol.~abs/1211.5063, 2012.

\bibitem{Martens2010}
J.~Martens, ``Deep learning via hessian-free optimization,'' in {\em ICML},
  2010.

\bibitem{Martens2011}
J.~Martens and I.~Sutskever, ``Learning recurrent neural networks with
  hessian-free optimization,'' in {\em Proceedings of the 28th international
  conference on machine learning (ICML-11)}, pp.~1033--1040, 2011.

\bibitem{Graves2013}
A.~Graves, ``Generating sequences with recurrent neural networks,'' {\em CoRR},
  vol.~abs/1308.0850, 2013.

\bibitem{Maass2002}
W.~{Maass}, T.~{Natschläger}, and H.~{Markram}, ``Real-time computing without
  stable states: A new framework for neural computation based on
  perturbations,'' {\em Neural Computation}, vol.~14, no.~11, pp.~2531--2560,
  2002.

\bibitem{Jaeger2001}
H.~Jaeger, ``The “echo state” approach to analysing and training recurrent
  neural networks,'' {\em GMD-Report 148, German National Research Institute
  for Computer Science}, 01 2001.

\bibitem{Laje2013}
R.~Laje and D.~Buonomano, ``Robust timing and motor patterns by taming chaos in
  recurrent neural networks,'' {\em Nature Neuroscience}, vol.~16, no.~7,
  pp.~925--935, 2013.

\bibitem{Triefenbach2010}
F.~Triefenbach, A.~Jalalvand, B.~Schrauwen, and J.-P. Martens, ``Phoneme
  recognition with large hierarchical reservoirs,'' in {\em Advances in Neural
  Information Processing Systems} (J.~Lafferty, C.~Williams, J.~Shawe-Taylor,
  R.~Zemel, and A.~Culotta, eds.), vol.~23, p.~9, Neural Information Processing
  System Foundation, 2010.

\bibitem{Vlachas2020}
P.~Vlachas, J.~Pathak, B.~Hunt, T.~Sapsis, M.~Girvan, E.~Ott, and
  P.~Koumoutsakos, ``Backpropagation algorithms and reservoir computing in
  recurrent neural networks for the forecasting of complex spatiotemporal
  dynamics,'' {\em Neural Networks}, vol.~126, pp.~191 -- 217, 2020.

\bibitem{Whittington2017}
J.~C.~R. Whittington and R.~Bogacz, ``An approximation of the error
  backpropagation algorithm in a predictive coding network with local hebbian
  synaptic plasticity,'' {\em Neural Computation}, vol.~29, no.~5,
  pp.~1229--1262, 2017.
\newblock PMID: 28333583.

\bibitem{Millidge2020}
B.~Millidge, A.~Tschantz, and C.~L. Buckley, ``Predictive coding approximates
  backprop along arbitrary computation graphs,'' 2020.

\bibitem{Walsh2020}
K.~S. Walsh, D.~P. McGovern, A.~Clark, and R.~G. O'Connell, ``Evaluating the
  neurophysiological evidence for predictive processing as a model of
  perception,'' {\em Annals of the New York Academy of Sciences}, vol.~1464,
  no.~1, pp.~242--268, 2020.

\bibitem{Dayan1995}
P.~Dayan, G.~E. Hinton, R.~M. Neal, and R.~S. Zemel, ``The helmholtz machine,''
  {\em Neural Computation}, vol.~7, no.~5, pp.~889--904, 1995.

\bibitem{Knill2004}
D.~C. Knill and A.~Pouget, ``The bayesian brain: the role of uncertainty in
  neural coding and computation,'' {\em Trends in Neurosciences}, vol.~27,
  no.~12, pp.~712 -- 719, 2004.

\bibitem{Ha2017}
D.~Ha and D.~Eck, ``A neural representation of sketch drawings,'' {\em CoRR},
  vol.~abs/1704.03477, 2017.

\bibitem{Philippsen2020}
A.~{Philippsen} and Y.~{Nagai}, ``A predictive coding account for cognition in
  human children and chimpanzees: A case study of drawing,'' {\em IEEE
  Transactions on Cognitive and Developmental Systems}, pp.~1--1, 2020.

\bibitem{PioLopez2016}
L.~Pio-Lopez, A.~Nizard, K.~Friston, and G.~Pezzulo, ``Active inference and
  robot control: a case study,'' {\em Journal of The Royal Society Interface},
  vol.~13, no.~122, p.~20160616, 2016.

\bibitem{Oliver2019}
G.~Oliver, P.~Lanillos, and G.~Cheng, ``Active inference body perception and
  action for humanoid robots,'' {\em CoRR}, vol.~abs/1906.03022, 2019.

\bibitem{Ciria2021}
A.~Ciria, G.~Schillaci, G.~Pezzulo, V.~V. Hafner, and B.~Lara, ``Predictive
  processing in cognitive robotics: a review,'' 2021.

\bibitem{Pitti2017}
A.~Pitti, P.~Gaussier, and M.~Quoy, ``Iterative free-energy optimization for
  recurrent neural networks (inferno),'' {\em PLOS ONE}, vol.~12, pp.~1--33, 03
  2017.

\bibitem{Annabi2020}
L.~{Annabi}, A.~{Pitti}, and M.~{Quoy}, ``Autonomous learning and chaining of
  motor primitives using the free energy principle,'' in {\em 2020
  International Joint Conference on Neural Networks (IJCNN)}, pp.~1--8, 2020.

\bibitem{Vinter2010}
A.~Vinter and E.~Chartrel, ``Effects of different types of learning on
  handwriting movements in young children,'' {\em Learning and Instruction},
  vol.~20, no.~6, pp.~476 -- 486, 2010.

\bibitem{Longcamp2005}
M.~Longcamp, M.-T. Zerbato-Poudou, and J.-L. Velay, ``The influence of writing
  practice on letter recognition in preschool children: A comparison between
  handwriting and typing,'' {\em Acta Psychologica}, vol.~119, no.~1, pp.~67 --
  79, 2005.

\bibitem{Taylor2009}
G.~W. Taylor and G.~E. Hinton, ``Factored conditional restricted boltzmann
  machines for modeling motion style,'' in {\em Proceedings of the 26th Annual
  International Conference on Machine Learning}, ICML '09, (New York, NY, USA),
  p.~1025–1032, Association for Computing Machinery, 2009.

\bibitem{Abrossimoff2018}
J.~{Abrossimoff}, A.~{Pitti}, and P.~{Gaussier}, ``Visual learning for reaching
  and body-schema with gain-field networks,'' in {\em 2018 Joint IEEE 8th
  International Conference on Development and Learning and Epigenetic Robotics
  (ICDL-EpiRob)}, pp.~197--203, 2018.

\bibitem{Lillicrap2016}
T.~P. Lillicrap, D.~Cownden, D.~B. Tweed, and C.~J. Akerman, ``Random synaptic
  feedback weights support error backpropagation for deep learning,'' {\em
  Nature communications}, vol.~7, p.~13276, November 2016.

\bibitem{Nokland2016}
A.~N{\o}kland, ``Direct feedback alignment provides learning in deep neural
  networks,'' in {\em Advances in Neural Information Processing Systems}
  (D.~Lee, M.~Sugiyama, U.~Luxburg, I.~Guyon, and R.~Garnett, eds.), vol.~29,
  pp.~1037--1045, Curran Associates, Inc., 2016.

\bibitem{Dua2019}
D.~Dua and C.~Graff, ``Uci machine learning repository,'' {\em
  [http://archive.ics.uci.edu/ml]. Irvine, CA: University of California, School
  of Information and Computer Science}, 2019.

\bibitem{Yamashita2008}
Y.~Yamashita and J.~Tani, ``Emergence of functional hierarchy in a multiple
  timescale neural network model: A humanoid robot experiment,'' {\em PLOS
  Computational Biology}, vol.~4, pp.~1--18, 11 2008.

\bibitem{Ijspeert2013}
A.~J. Ijspeert, J.~Nakanishi, H.~Hoffmann, P.~Pastor, and S.~Schaal,
  ``Dynamical movement primitives: Learning attractor models for motor
  behaviors,'' {\em Neural Computation}, vol.~25, no.~2, pp.~328--373, 2013.
\newblock PMID: 23148415.

\bibitem{Braud2018}
R.~{Braud}, A.~{Pitti}, and P.~{Gaussier}, ``A modular dynamic sensorimotor
  model for affordances learning, sequences planning, and tool-use,'' {\em IEEE
  Transactions on Cognitive and Developmental Systems}, vol.~10, no.~1,
  pp.~72--87, 2018.

\bibitem{Tsuda2015}
I.~Tsuda, ``Chaotic itinerancy and its roles in cognitive neurodynamics,'' {\em
  Current Opinion in Neurobiology}, vol.~31, pp.~67 -- 71, 2015.
\newblock SI: Brain rhythms and dynamic coordination.

\bibitem{Inoue2020}
K.~Inoue, K.~Nakajima, and Y.~Kuniyoshi, ``Designing spontaneous behavioral
  switching via chaotic itinerancy,'' 2020.

\end{thebibliography}

\newpage
\onecolumn
\begin{appendices}
\section{MTRNN training procedure}
\label{annex:mtrnn_algo}

The training procedure for the MTRNN described in \cite{Mochizuki2013} comprises two phases: a motor babbling phase and an imitation learning phase. The motor babbling phase allows for a proper initialization of the MTRNN weights using sensorimotor trajectories collected with random interactions with the environment. The imitation learning phase trains the model in order to generate desired trajectories with a supervision in the sensory (here visual) space. Here is how we implemented these two phases:

\begin{algorithm}
    \SetAlgoLined
    Initialize the MTRNN \;
    \For{$0 \leq i < I_1$}{
        Generate a motor trajectory with a random initial hidden state $\mathbf{h}_0$\;
        Perform the motor trajectory $(\mathbf{m}_t)_{0\leq t < T}$\;
        Collect the corresponding visual trajectory $(\mathbf{o}_t)_{0\leq t < T}$\;
        Train the MTRNN using the visual trajectory as target and the same initial hidden state $\mathbf{h}_0$\;
    }
    \caption{Motor babbling phase}
\end{algorithm}

\begin{algorithm}
    \SetAlgoLined
    \For{$0 \leq i < I_2$}{
        \For{$0 \leq k < p$}{
            Infer through BPTT the initial hidden state $\mathbf{h}_{0}$ of the MTRNN generating the target visual trajectory $(\mathbf{o}^*_{t, k})_{0\leq t < T}$\;
            Generate the motor trajectory with the obtained initial hidden state $\mathbf{h}_{0}$\;
            Perform the motor trajectory $(\mathbf{m}_{t})_{0\leq t < T}$\;
            Collect the corresponding visual trajectory $(\mathbf{o}_{t})_{0\leq t < T}$\;
            Train the MTRNN using the visual trajectory as target and the same initial hidden state $\mathbf{h}_{0}$\;
        }
    }
    \caption{Imitation learning phase}
\end{algorithm}

In these algorithms, $I_1$ and $I_2$ respectively denote the number of training iterations in the babbling phase and in the imitation phase, and $p$ denotes the number of visual trajectories in the training data set that we learn to reproduce. In our experiments, we varied $p$, and used $I_1=3000$ and $I_2=2000$.

Note that in the MTRNN model, there are actually several hidden states corresponding to the different timescales of the model. In our experiments, we used a model with two timescales $\tau_f = 5$ (fast) and $\tau_s = 10$ (slow), with hidden state dimensions of $n_f = n_s = n/2$, where $n$ is the total hidden state dimension.

\section{Repository}

The code for the implementation and training of our model is made available in the following repository: \url{https://github.com/sino7/bidirectional-interaction-between-visual-and-motor-generative-models}.

\section{Examples of generated trajectories}
\label{annex:trajectory_examples}

In this appendix, we display more examples of trajectories obtained in our experiments with perturbations on the motor output, transformations of the visual predictions, and impairments on the visual RNN.

\begin{figure*}[!ht]
    \centering
    \includegraphics[width=\textwidth]{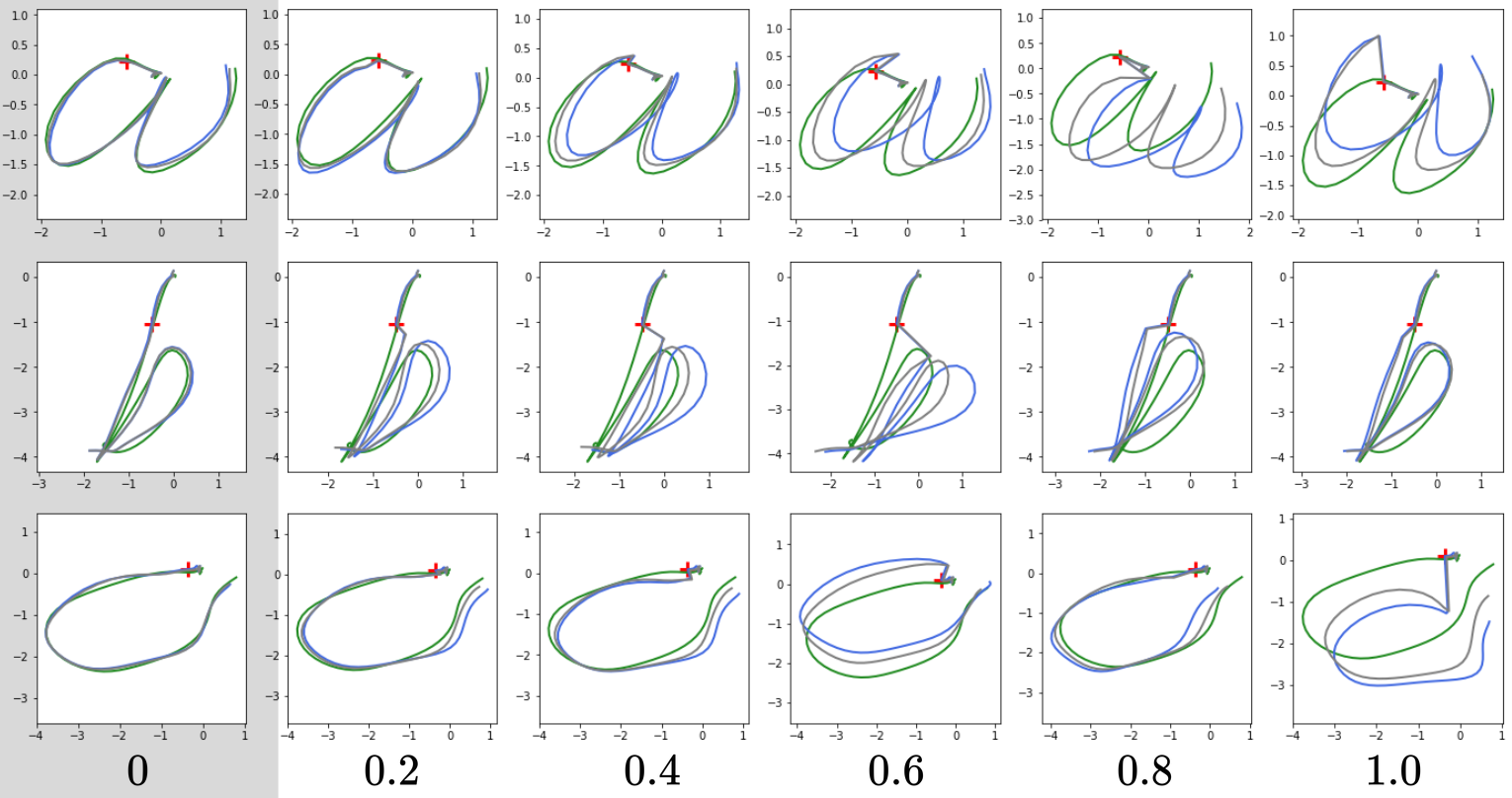}
    \caption{Examples of trajectories obtained in the perturbation experiment, for different perturbation amplitudes $\sigma_p^2 \in \{0.0, 0.2, 0.4, 0.6, 0.8, 1.0\}$. Blue lines represent the uncontrolled motor trajectories displayed in the visual space. Green lines represent the predicted visual trajectories. Gray lines represent the corrected motor trajectories displayed in the visual space. These trajectories were obtained with an RNN hidden state dimension of 50 after training.}
    \label{fig:examples_perturbations}
\end{figure*}

\begin{figure*}[!ht]
    \centering
    \includegraphics[width=\textwidth]{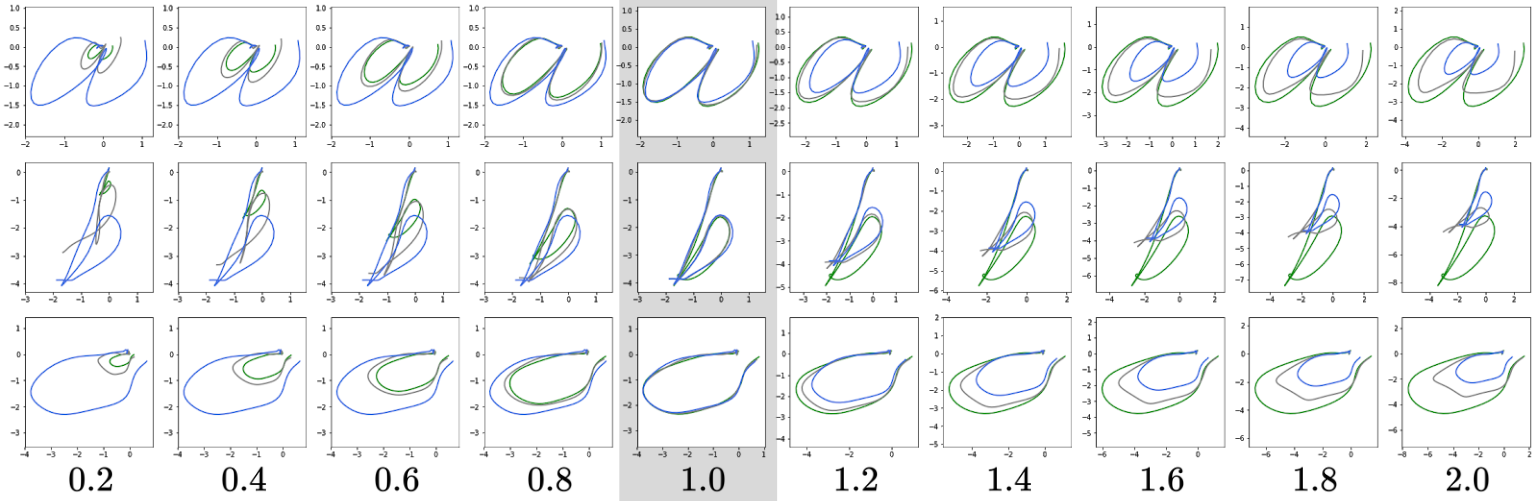}
    \caption{Examples of trajectories obtained in the scaling experiment, for different scales $s \in \{0.2, 0.4, 0.6, 0.8, 1.0, 1.2, 1.4, 1.6, 1.8, 2.0\}$. Blue lines represent the uncontrolled motor trajectories displayed in the visual space. Green lines represent the rescaled visual trajectories. Gray lines represent the corrected motor trajectories displayed in the visual space. These trajectories were obtained with an RNN hidden state dimension of 50 after training.}
    \label{fig:examples_scaling}
\end{figure*}

\begin{figure*}[!ht]
    \centering
    \includegraphics[width=\textwidth]{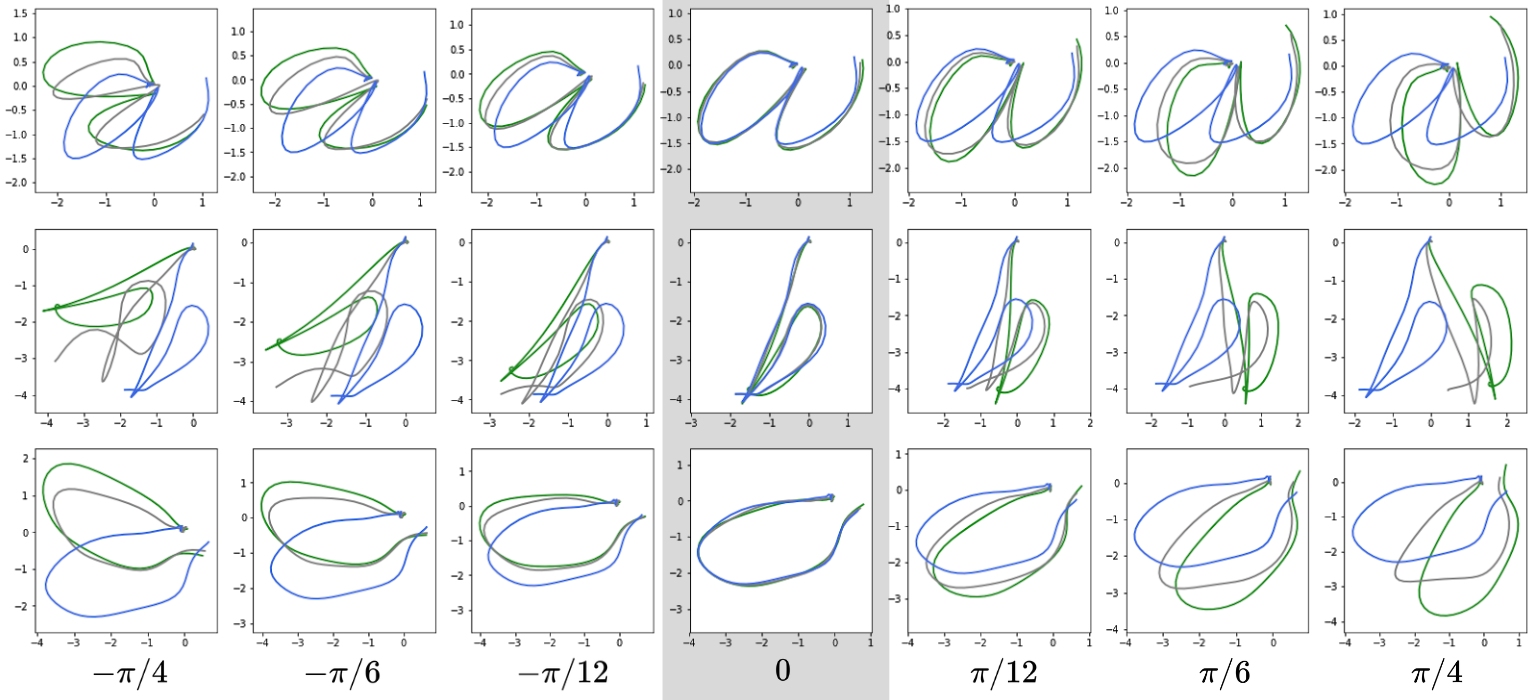}
    \caption{Examples of trajectories obtained in the rotation experiment, for different rotation angles $\theta \in \{-\pi/4, -\pi/6, -\pi/12, 0, \pi/12, \pi/6, \pi/4\}$. Blue lines represent the uncontrolled motor trajectories displayed in the visual space. Green lines represent the rotated visual trajectories. Gray lines represent the corrected motor trajectories displayed in the visual space. These trajectories were obtained with an RNN hidden state dimension of 50 after training.}
    \label{fig:examples_rotation}
\end{figure*}

\begin{figure*}[!ht]
    \centering
    \includegraphics[width=\textwidth]{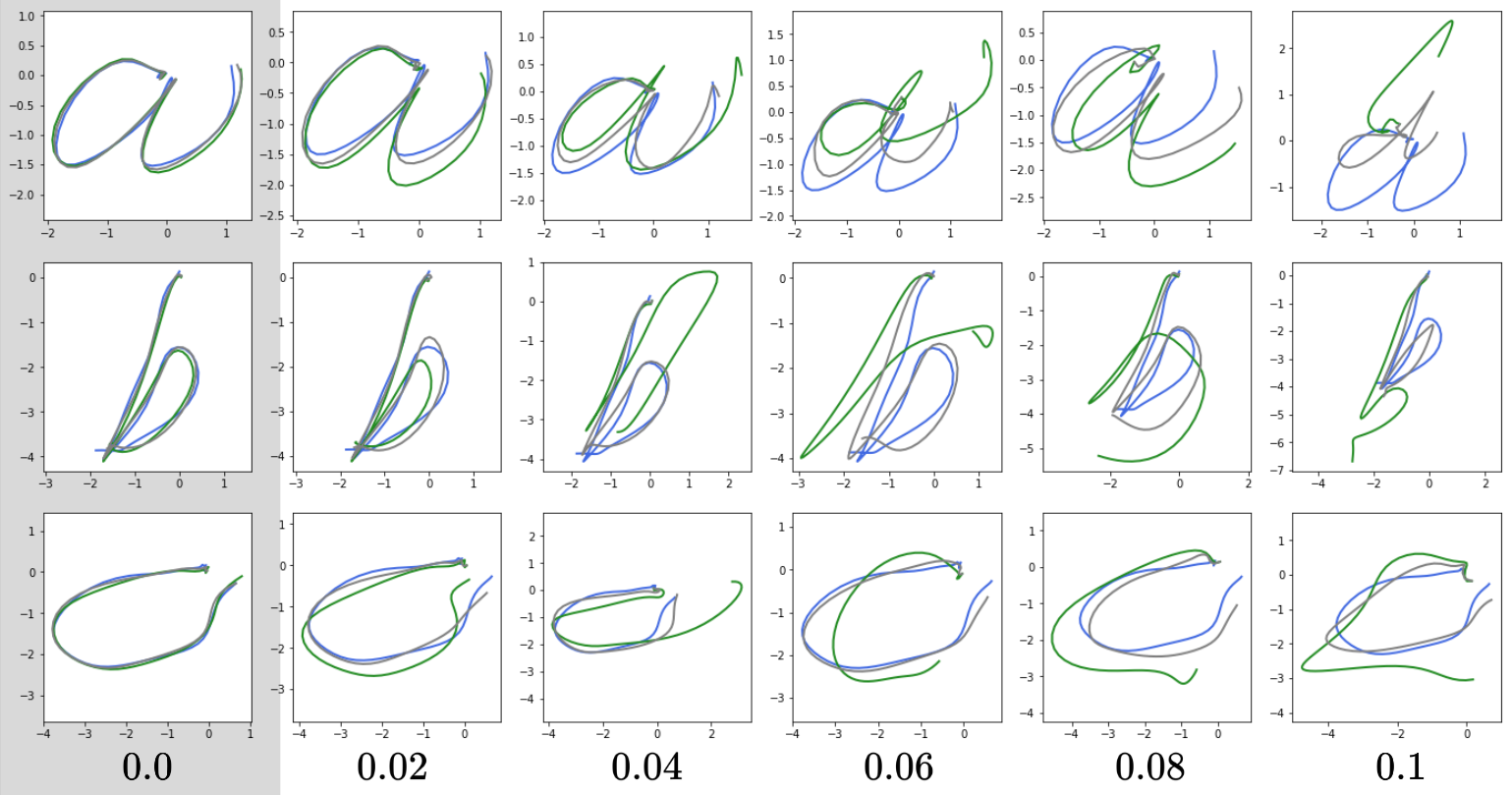}
    \caption{Examples of trajectories obtained in the visual impairment experiment, for different impairment amplitudes $\sigma_i^2 \in \{0.0, 0.02, 0.04, 0.06, 0.08, 0.1\}$. Blue lines represent the uncontrolled motor trajectories displayed in the visual space. Green lines represent the visual trajectories predicted by the impaired visual RNN. Gray lines represent the corrected visual trajectories. These trajectories were obtained with an RNN hidden state dimension of 50 after training.}
    \label{fig:examples_impairments}
\end{figure*}

\end{appendices}

\end{document}